\documentclass[conference]{IEEEtran}
\IEEEoverridecommandlockouts

\usepackage[mode=buildnew]{standalone}
\usepackage{amssymb}
\usepackage{cite}
\usepackage{amsmath,amssymb,amsfonts}
\usepackage{algorithmic}
\usepackage{algorithm}
\usepackage{graphicx}
\usepackage{textcomp}
\usepackage{xcolor}
\usepackage[table]{xcolor}
\usepackage{caption}
\usepackage{pifont}
\usepackage{orcidlink}
\usepackage{array}

\def\BibTeX{{\rm B\kern-.05em{\sc i\kern-.025em b}\kern-.08em
    T\kern-.1667em\lower.7ex\hbox{E}\kern-.125emX}}  

\begin{document}

\title{Agentic AI-Based Joint Computing and Networking via Mixture of Experts and Large Language Models}

\author{Robert-Jeron Reifert\,\orcidlink{0000-0003-3922-8996},~\IEEEmembership{Graduate Student Member,~IEEE,}
    Alaa Alameer Ahmad\,\orcidlink{0000-0002-0764-5560},\\
	Hayssam Dahrouj\,\orcidlink{0000-0002-0737-6372},~\IEEEmembership{Senior Member,~IEEE,} and
	Aydin Sezgin\,\orcidlink{0000-0003-3511-2662},~\IEEEmembership{Senior Member,~IEEE}
\thanks{
	This work has been submitted to the IEEE for possible publication. Copyright may be transferred without notice, after which this version may no longer be accessible.\newline
	This work is supported in part by the German Federal Ministry of Research Technology and Space (BMFTR) in the course of the 6GEM+ Transfer Hub under grant 16KIS2411. (\emph{Corresponding author: Robert-Jeron Reifert})\newline
	Robert-Jeron Reifert and Aydin Sezgin are with Digital Communication Systems, Ruhr University Bochum, Bochum, Germany. (\{robert-.reifert,aydin.sezgin\}@rub.de). 
	Hayssam Dahrouj is with the Department of Electrical Engineering, University of Sharjah, Sharjah, United Arab Emirates. (hayssam.dahrouj@gmail.com). 
	Alaa Alameer Ahmad is with Cariad SE, Wolfsburg, Germany. (alaa.alameer85@gmail.com).}}

\maketitle

\begin{abstract}
Future sixth-generation (6G) mobile networks are envisioned to be equipped with a diverse set of powerful, yet highly specialized, optimization experts. Such a promising vision is concurrently expected to give rise to the need for scalable mechanisms that can select, combine, and orchestrate such experts based on high-level intent and uncertainty descriptions. In this paper, we propose an agentic artificial intelligence (AI)-based network optimization framework that integrates mixture of experts (MoE) architectures with large language models (LLMs). Under the proposed framework, the employed LLM acts as a semantic gate to reason over operator objectives and dynamically compose suitable optimization agents. 
The proposed framework is formulated in a model-agnostic manner and bridges human-readable network intents with low-level resource allocation decisions, enabling flexible optimization across heterogeneous objectives and operating conditions. As a representative instantiation, we apply the framework to a joint communication and computing network and design a library of specialized optimization experts covering throughput, fairness, and delay-driven objectives under both regular and robust conditions. Numerical simulations demonstrate that the proposed agentic MoE framework consistently achieves near-optimal performance compared to exhaustive expert combinations while outperforming individual experts across diverse objectives, including delay minimization and throughput maximization.
\end{abstract}


\section{Introduction}
The integration of artificial intelligence (AI), particularly agentic and generative AI, into the design and optimization of mobile networks has attracted significant attention from both academia and industry. This trend is especially prominent in the context of emerging sixth-generation (6G) wireless networks and beyond \cite{11097898, chatzistefanidis2025mx}. Over the past generations, mobile wireless networks have evolved from relatively static communication infrastructures into highly dynamic and heterogeneous systems that jointly manage communication, computing, and control across multiple layers and domains \cite{8869705}. In particular, the emerging 6G architectures are poised to incorporate remote data processing, task offloading, and edge computing capabilities, further emphasizing the tight coupling between computing and networking functionalities \cite{10258360}.  For example, in edge-assisted extended reality (XR) applications, video frames must be rendered or processed at the base station (BS) before being transmitted to user devices \cite{10666455}. Unlike previous wireless network generations, 6G networks are envisioned to be \emph{AI-native}, where intelligence is inherently intrinsic to the network architecture and operation \cite{8808168}.

While AI-native networks promise unprecedented adaptability, they concurrently spearhead a vast landscape of AI- and learning-based optimization methods, spanning robust resource allocation, fairness-aware scheduling, latency minimization, and joint communication and computing optimization \cite{10071987,reifert2024robust}. Despite their mutual interdependence, existing optimization approaches remain highly specialized, with each method tailored to a specific objective, system assumption, or uncertainty model, and are often difficult to reconcile within a unified optimization framework.

Further, the increasing complexity of radio access networks, driven by evolving specifications, large volumes of heterogeneous data, and rapidly changing operating environments, motivates the need for agentic AI in future wireless systems \cite{chatzimiltis2025agentic,11162291, tong2026wirelessagentautomatedagenticworkflow}. Unlike conventional closed-loop or passive learning frameworks, agentic AI enables autonomous decision-making, self-organization, and dynamic adaptation by actively reasoning over system states, objectives, and available actions \cite{xiao2025towards}.


Building onto the above advancement, recent works \cite{10929033, 10579547} envision \emph{artificial general intelligence (AGI)-native networks}, in which AGI-like capabilities enable networks to reason across heterogeneous tasks, coordinate diverse optimization mechanisms, and autonomously manage resources. In such architectures, the network must not only optimize individual functions but also orchestrate multiple specialized capabilities in response to dynamic objectives and environmental conditions.

Within such an agentic AI-based framework, mixture of experts (MoE) architectures offer a natural mathematical framework for combining multiple specialized solutions \cite{6215056}. However, conventional gating mechanisms are typically trained for narrow tasks and operate on low-level features, which limits their ability to interpret high-level intent, natural-language objectives, or abstract system-level trade-offs \cite{11284842}.

Recent advances in large language models (LLMs) provide a practical and scalable means to realize agentic intelligence in network optimization, owing to their strong capabilities in semantic reasoning, intent interpretation, and tool orchestration \cite{10685369}. These properties make LLMs particularly attractive as agentic gate networks, capable of mapping high-level, human-readable optimization goals to structured combinations of expert solutions \cite{11284842}.

Motivated by these observations, this paper proposes a generic network optimization framework based on an LLM-enabled MoE architecture, as illustrated in Fig.~\ref{fig:framdl}, with a special focus on joint computing and networking systems optimization. Specifically, we consider the coordinated allocation of wireless communication resources and edge computing resources, including transmit power, computing power, and processing cycle assignments, which together determine the achievable network throughput, fairness, latency, and task processing performance. The framework leverages an agentic LLM to select and weight optimization agents based on high-level network objectives and uncertainty descriptions. In this work, an \emph{expert} refers to a specialized optimization agent that determines resource allocations from the observed network state. Such experts may be implemented using model-based or learning-based methods, e.g., deep learning optimization networks. The LLM acts as a high-level reasoning layer that interprets operator intent and orchestrates suitable experts. This architecture reflects emerging agentic and AGI-inspired systems, where a central reasoning model coordinates multiple specialized capabilities to solve complex tasks. 

\subsection{Related Works}
The integration of LLMs, MoE architectures, and wireless network optimization has recently attracted increasing research attention. 
Several recent studies have explored the use of MoE architectures with LLM-based gating mechanisms. For instance, in \cite{du2024mixture,10592370}, an MoE framework is proposed in which an LLM oversees and selects specialized experts based on user requests, while deep reinforcement learning is employed to guide expert selection. Although conceptually related to our work, these approaches primarily focus on single-user scenarios and simplified environments, without considering network-wide resource coupling, optimization constraints, or joint communication and computing aspects.
A comprehensive overview of MoE architectures in wireless systems is provided in \cite{11284842}, which surveys decentralized generative AI and reinforcement learning approaches and discusses LLM-based gating as an emerging direction. 

An alternative research direction investigates the use of pretrained LLMs as optimization or control agents. In \cite{lee2025convergence}, LLMs are employed directly as network optimization agents through iterative prompt engineering and evolutionary-style search mechanisms. Similarly, \cite{11173862} integrates LLM-based optimization with classical iterative algorithms, using LLMs to update subsets of optimization variables in each iteration. In \cite{wang2026wirelesspowercontrolbased}, an LLM within a physics-informed framework is used in the context of wireless power control. While these approaches demonstrate the feasibility of LLM-driven optimization, they treat LLMs as direct optimizers, which can incur substantial computational overhead and do not leverage specialized domain-specific optimization experts.


Several works have investigated the role of foundation models and LLMs in future wireless architectures. For instance, the authors of \cite{11103462} propose an AI-native framework that integrates foundation models, expert knowledge, and LLM-based reasoning modules for future 6G networks. 
Other studies focus on enabling user interaction and network programmability through LLMs. For instance, \cite{10850042} introduces a network-oriented LLM interface that enables script-free interaction with network simulators, while \cite{11097898} proposes TelecomGPT, a domain-adapted LLM enriched with wireless communication knowledge. These works primarily address usability and knowledge representation rather than network optimization.

Beyond optimization, LLMs have been explored in broader network operations and management contexts. The survey in \cite{liu2025large} provides a comprehensive overview of LLM applications in network management, orchestration, and automation, including early discussions of expert-based reasoning paradigms. Related directions such as knowledge distillation toward the network edge are studied in \cite{wu2025towards}. 

In contrast to prior studies, the current paper proposes a mathematically grounded, system-level framework that integrates MoE optimization with agentic LLM-based reasoning. The proposed approach differs from existing work in three key aspects: 
\begin{itemize}
    \item[$i)$] Optimization-driven design, by explicitly formulating network optimization objectives, constraints, and robustness requirements; 
    \item[$ii)$] Agentic gating, where the LLM is employed solely as a semantic gate to select and weight specialized experts; 
    \item[$iii)$] Joint communication and computing perspective, capturing a network-wide coupling between communication and computing resources.
\end{itemize}
To the best of our knowledge, this is the first work in the existing literature which proposes such an integrated MoE-LLM-based framework in the context of joint computing and networking optimization. Details on the contributions of the work are discussed next.

\subsection{Contributions}
\begin{figure*}[t]
\centering
\includegraphics[width=.8\linewidth]{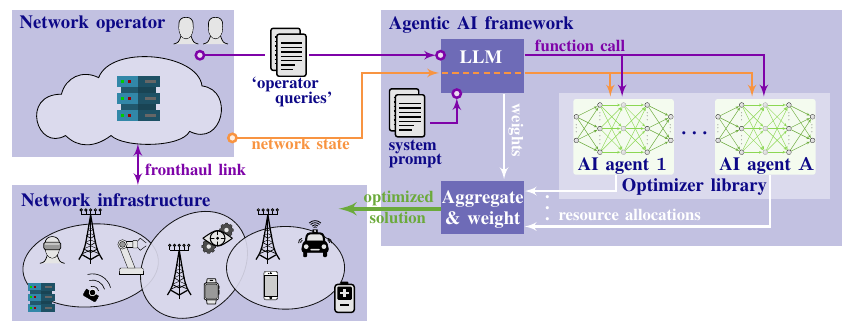}\vspace*{-.2cm}
\caption{General agentic AI-based network optimization framework with an LLM-enabled gate and a library of deep learning-based optimization experts, illustrating the interaction between network operator intent and network infrastructure.}
\label{fig:framdl}
\vspace*{-.4cm}
\end{figure*}%
This paper introduces a novel agentic network optimization paradigm that combines MoE architectures with LLMs. By leveraging an LLM as a high-level gate, the proposed framework enables semantic reasoning over network objectives and uncertainty descriptions, allowing the dynamic selection and composition of specialized optimization experts. In doing so, the framework bridges human-readable operator intent and low-level network resource allocation decisions, providing a flexible and scalable foundation for joint communication and computing optimization in AI-native networks. In context of the paper, the adopted joint optimization framework refers to the coordinated allocation of wireless communication resources and edge computing resources, namely, transmit power, computing power, and computing cycle assignments for task processing.
The main contributions of this work are summarized as follows:
\begin{itemize}
    \item \textbf{Agentic LLM-driven MoE framework:}  
    We propose a generic, model-agnostic network optimization framework in which an LLM acts as an agentic gate to select and weight heterogeneous optimization agents. The LLM takes as input a system prompt describing the available agents, the current network state, and a human-readable operator query specifying the optimization objective, and outputs a structured selection and weighting of agents.

    \item \textbf{Unified expert-based formulation:}  
    A mathematical framework is developed that captures diverse objectives, constraints, and robustness requirements within a MoE paradigm.

    \item \textbf{Joint communication and computing case study:}  
    The framework is instantiated for a joint communication–computing system, including coupled power budgets, delay models, and uncertainty-aware optimization.

    \item \textbf{Comprehensive expert library:}  
    We design and evaluate $30$ specialized experts covering throughput-, fairness-, and delay-driven objectives under both regular and robust network conditions.   

    \item \textbf{Numerical validation:} 
    Extensive simulations demonstrate that the proposed agentic AI-based framework achieves near-optimal performance, high feasibility and accuracy, and competitive results compared to exhaustive search, while avoiding its exponential complexity.
    
\end{itemize}

The remainder of this paper is organized as follows. Section~\ref{sec:2} presents a general discussion of the proposed network optimization framework based on agentic AI, LLMs, and MoE. Section~\ref{sec:cs} addresses the details of the joint communication and computing optimization case study using the proposed framework, including the system model and objective formulation (Sec.~\ref{ssec:A}), the design of deep learning-based experts (Sec.~\ref{ssec:B}), and an example agent (Sec.~\ref{ssec:C}). The corresponding simulation results and insights are discussed in Section~\ref{sec:cs2}. Finally, Section~\ref{sec:con} concludes the paper.

\section{Generic Network Optimization Framework via Agentic AI}\label{sec:2}%
This section presents a generic network optimization framework based on an LLM-enabled MoE architecture. The framework is intentionally formulated at an abstract level and does not assume a specific communication or computing model. The section instead captures the logical structure of agentic decision-making for network optimization, enabling the dynamic selection and composition of specialized optimization experts. Concrete system models and objective functions are introduced later in Section~\ref{sec:cs} as a representative case study.

Fig.~\ref{fig:framdl} illustrates the overall proposed agentic AI-based framework, which proceeds from an abstract network state representation to expert-based optimization and LLM-driven orchestration. The figure depicts how network operator intent and network state information are provided to the agentic AI-based framework, where an LLM-enabled gate processes operator queries together with a system prompt. Based on this input, the LLM selects and orchestrates suitable AI-based optimization agents from a library of optimizers. The resulting expert outputs are then aggregated and weighted according to the inferred gating decisions, yielding a final optimized solution that is applied to the network infrastructure. In the following, we present the individual components and logical flow of Fig.~\ref{fig:framdl} in detail.


\subsection{Network Optimization}
We consider a generic network optimization problem characterized by an abstract network state $\mathcal{S}$, which captures all relevant information about the current operating conditions, such as channel states, traffic demands, computing workloads, and uncertainty descriptions. Based on $\mathcal{S}$, the network aims to optimize a utility function subject to system constraints.

Let $\mathbf{x}$ denote a vector of optimization variables, which may represent, for example, communication resources, computing allocations, or joint control parameters. The generic optimization problem is formulated as
\begin{subequations}\label{eq:Opt0}
\begin{align}
	\underset{\mathbf{x}}{\text{max}}\quad &f_{\text{utility}}(\mathbf{x})\tag{\ref{eq:Opt0}} \\
    \text{s.t.} \quad\, & \mathbf{x}\in\mathcal{C}(\mathbf{x}),
\end{align}
\end{subequations}
where $f_{\text{utility}}(\mathbf{x})$ denotes a specific performance objective (e.g., throughput, fairness, or latency), and $\mathcal{C}(\mathbf{x})$ represents the set of feasibility constraints imposed by physical, computational, or quality-of-service limitations.

This formulation is intentionally kept abstract to accommodate a wide range of network optimization objectives and system models.

\subsection{Optimization Agents}
Solving \eqref{eq:Opt0} from scratch for every network state is generally challenging not only due to computational complexity, but also because the optimization problem itself may vary across operating scenarios. In practical networks, the objective function, constraint set, and the set of optimization variables can change dynamically depending on the service requirements, system configuration, and uncertainty conditions. Consequently, no single solver or optimization algorithm is sufficient to address all possible problem instances.

To address this challenge, the paper assumes access to a library of $A$ specialized optimization agents, where each agent is designed to solve a particular class of network optimization problems or objectives. Let
\begin{equation}
    \Phi = \{\mathcal{A}_1, \ldots, \mathcal{A}_A\}
\end{equation}
denote the set of available agents, where each expert $\mathcal{A}_a$ may correspond to a deep learning-based optimizer, a model-driven algorithm, or a hybrid learning-optimization method, and $a \in \Phi$ indexes the agents\footnote{In this paper, the terms \emph{agent}, \emph{expert}, and \emph{optimizer} are used interchangeably, viz., a specialized algorithm that determines network resource allocations.}.

Given the network state $\mathcal{S}$, agent $\mathcal{A}_a$ produces a candidate solution
\begin{equation}
    \mathbf{x}_a \in \mathbb{R}^{d_a^\text{out}},
\end{equation}
where $d_a^\text{out}$ denotes the dimensionality of the output of agent $a$. These expert solutions may correspond to different optimization goals, such as throughput maximization, fairness-aware allocation, robust optimization, or latency minimization. Such distinct optimization perspectives are captured in the abstract utility functions
\begin{equation}
    f_{\text{utility},a}(\mathbf{x}_a), \quad \forall a\in\Phi,
\end{equation}
where $f_{\text{utility},a}(\mathbf{x}_a)$ refers to agent $a$'s utility function.

While the proposed framework does not restrict the internal structure of the experts, the case study in Section~\ref{sec:cs} focuses on deep learning-based implementations due to their suitability for real-time inference. Further, each expert operates independently and does not account for the presence of other experts, resulting in a diverse but fragmented set of candidate solutions.

\subsection{LLM-enabled Gate Network}
The MoE paradigm provides a principled mechanism for combining specialized optimization models. However, selecting and weighting suitable experts from $\Phi$ requires reasoning over high-level objectives, system context, and uncertainty descriptions, which is beyond the scope of conventional gating functions.

To address this challenge, we adopt an agentic AI approach. That is, in the proposed framework, an LLM acts as an agentic gate that interprets network intent and context, and determines both expert selection and combination. This results in an LLM-enabled gate network that serves as the decision-making layer of the MoE architecture.

The gate then receives two types of inputs:
\begin{itemize}
	\item A system prompt $\mathcal{Q}^\text{sys}$, which encodes global rules, constraints, as well as task and expert descriptions.
	\item An operator query $\mathcal{Q}^\text{op}(\mathcal{S})$, which expresses high-level optimization intent in natural language based on the current network state $\mathcal{S}$.
\end{itemize}
Exemplary high-level inputs to the system prompt and operator query are provided in Table~\ref{tab:abstract_prompts}. The system prompt first establishes the LLM's role within the framework as a gating mechanism, followed by a description of the available agents, including their utility functions, objectives, and operating conditions. The operator query specifies the network setup (e.g., constraints and system parameters) and the desired objective, or a combination thereof.

\begin{table}[t]
\caption{High-level illustration of system prompt $\mathcal{Q}^{\text{sys}}$ and operator query $\mathcal{Q}^\text{op}(\mathcal{S})$.}
\label{tab:abstract_prompts}
\centering
\small
\begin{tabular}{r p{0.9\linewidth}}
        \multicolumn{2}{l}{\cellcolor{black!50}\textcolor{white}{\textbf{System prompt}}}\\
        \cellcolor{black!50}\textcolor{white}{} & \emph{You are functioning as a gate network that receives a query from the network operator and your task is to route the question to a subset of optimization expert from a library. You have $A$ experts that can be used to resolve queries either by their own or on combinations. This depends on the query if the requested information need to be a combination of the results from several experts or one expert can fully address the query. This can be solely determined by the description of each expert area of expertise. Here is a detailed description of the available experts and their area of specialization:}\\
        \cellcolor{black!50}\textcolor{white}{} & \cellcolor{black!10}\emph{-Agent description (including utility function, goal, operating conditions, parameter set, etc.)-}\\
        \multicolumn{2}{l}{\cellcolor{black!50}\textcolor{white}{\textbf{Operator query}}}\\
        \cellcolor{black!50}\textcolor{white}{} & \emph{-Network setup description (e.g., XR, computing, imperfect channels, etc.)-}\\
        \cellcolor{black!50}\textcolor{white}{} & \cellcolor{black!10}\emph{-Operator goal description (e.g., sum-rate maximization, etc.)-}
    \end{tabular}
\end{table}

Based on these inputs, the LLM produces:
\begin{itemize}
	\item a binary selection vector $\mathbf{a} \in \{0,1\}^{A}$, indicating which experts are activated,
	\item a continuous weighting vector $\boldsymbol{\alpha} = [\alpha_1,\ldots,\alpha_A]^T$, where $\alpha_a \geq 0$ and $\sum_{a} \alpha_a = 1$.
\end{itemize}

This mapping is expressed as
\begin{equation}\label{eq:LLM}
    (\mathbf{a}, \boldsymbol{\alpha}) = \mathcal{G}_{\text{LLM}}\big(\mathcal{Q}^\text{sys}, \mathcal{Q}^\text{op}(\mathcal{S})\big),
\end{equation}
where $\mathcal{G}_{\text{LLM}}(\cdot)$ denotes the LLM-based gating function.

The final network control decision is obtained by composing the selected expert solutions according to
\begin{equation}\label{eq:combexp}
    \mathbf{x} = \sum_{a \in \Phi} \alpha_a \mathbf{x}_a,
\end{equation}
where $\alpha_a = 0$ for experts that are not selected, and $\alpha_a=1$ otherwise.

The proposed framework enables scalable network optimization by decomposing a complex decision-making problem into a set of specialized deep learning-based experts and an agentic LLM-based gate. Rather than exhaustively searching over all possible expert combinations and parameterizations, the LLM infers suitable expert selections and mixture weights directly from high-level network intent and state descriptions. This abstraction allows the framework to operate efficiently in real time while remaining agnostic to specific system models, thereby facilitating its application across diverse communication and computing scenarios.

\section{Case Study: Joint Communication and Computing Optimization}\label{sec:cs}
Based on the general framework proposed above, we now illustrate the prospects of such agentic AI-based optimization approach in the case of joint computing and communication systems optimization. The considered case study is organized into three parts. First, the joint communication and computing system model is discussed, followed by a presentation of network optimization objectives. We then briefly discuss the proposed deep learning-based experts and the LLM-based aggregation and weighting. Finally, for the sake of illustration, we provide an example of one particular optimization problem and its corresponding AI expert formulation.

\subsection{System Model and Network Objectives}\label{ssec:A}
\begin{figure}[t]
\centering
\includegraphics[width=.9\linewidth]{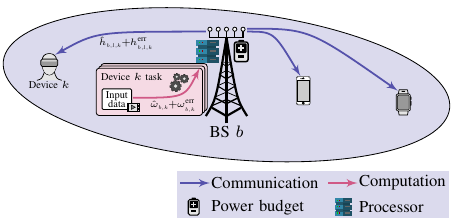}\vspace*{-.2cm}
\caption{Joint communication and computing network consisting of a multi-antenna BS with embedded computing capabilities serving multiple single-antenna devices.}
\label{fig:sysmdl}
\vspace*{-.4cm}
\end{figure}
In this section, we consider a generic joint communication and computing network model, applicable to emerging use cases such as digital twinning \cite{11023535} and XR \cite{10328648}. The network comprises a single BS equipped with embedded computing capabilities, serving a total of $K$ single-antenna devices. The BS is assumed to be equipped with $L$ antennas, while the set of devices is denoted by $\mathcal{K}=\{1,\ldots,K\}$.

Fig.~\ref{fig:sysmdl} illustrates the considered system model, highlighting both communication and computing components \cite{reifert2024robust}. In the following, we introduce the system assumptions, signal model, and resource constraints. For later use in the expert-based optimization framework, the derived constraints and key equations are collected into structured sets, enabling their flexible reuse across different optimization objectives and expert formulations. We begin by describing the communication network model.

\subsubsection{Networking Model}
Let $h_{l,k}\in\mathbb{C}$ denote the channel coefficient from the $l$-th antenna of the BS to device $k$. The aggregate channel vector of device $k$ is given by $\mathbf{h}_{k} = [h_{1,k},\cdots,h_{L,k}]^T$. The beamforming vector associated with device $k$ is denoted by $\mathbf{v}_{k}\in\mathbb{C}^{L}$ and defined as $\mathbf{v}_{k} = [v_{1,k},\cdots,v_{L,k}]^T$, where $\lVert \mathbf{v}_{k} \rVert_2 = 1$. Let $p_k^{\text{tx}}$ denote the transmit power allocated to device $k$, and define the transmit power vector as $\mathbf{p}^\text{tx} = [p_{1}^\text{tx},\cdots,p_{K}^\text{tx}]^T$, where the superscript `$\text{tx}$' denotes transmission-related variables.

The transmitted message intended for device $k$ is denoted by $\xi_k$, for $k=1,2,\ldots, K$, where $\xi_k \sim \mathcal{CN}(0,1)$ are independent and identically distributed circularly symmetric complex Gaussian random variables. The total transmit power constraint at the BS is expressed as
\begin{align}
    \mathcal{P}_\text{comm}:\quad &\sideset{}{_{k\in\mathcal{K}}}\sum p_k^\text{tx} \leq P^\text{max,tx},\label{eq:Pcomm}
\end{align}
where $P^{\text{max,tx}}$ denotes the maximum available transmit power. Each device receives its desired signal, interference from transmissions intended for other devices, and additive noise. Accordingly, the received signal at device $k$ is given by
\begin{align}
y_k = \mathbf{h}_{k}^H \mathbf{v}_{k} \sqrt{p^\text{tx}_k} \xi_k + \sideset{}{_{j\in\mathcal{K}\backslash\{k\}}}{\sum} \mathbf{h}_{k}^H \mathbf{v}_{j} \sqrt{p^\text{tx}_j} \xi_j + n_k, \label{eq:yk}
\end{align}
where $n_k$ denotes additive white Gaussian noise with zero mean and variance $\sigma^2$.
The signal-to-interference-plus-noise ratio (SINR) of device $k$ and its associated achievable rate are expressed as
\begin{align}
\mathcal{D}_\text{comm}:\quad &r_k^\text{tx} = W\log_2\left(1+\Gamma_k\right), &\forall k\in\mathcal{K},\label{eq:rk}\\
&\Gamma_k = \frac{|\mathbf{h}_k^H \mathbf{v}_k|^2 p^\text{tx}_k}{\sum_{j\in\mathcal{K}\backslash\{k\}} |\mathbf{h}_k^H \mathbf{v}_j|^2 p^\text{tx}_j + \sigma^2}, \hspace*{-.1cm} &\forall k\in\mathcal{K},\label{eq:Gammak}
\end{align}
where $W$ denotes the system bandwidth. The above expressions are collected in the constraint set $\mathcal{D}_{\text{comm}}$. Given the output data size $D_k^{\text{out}}$ of the computing task associated with device $k$, the resulting communication delays are expressed as
\begin{align}
    \mathcal{T}_\text{comm}:\quad &t_k^\text{tx} = \frac{D_k^\text{out}}{r_k^\text{tx}}, &&\forall k\in\mathcal{K}.\label{eq:tcomm}
\end{align}
In analogy to the networking model, we next describe the computing model in details.

\subsubsection{Computing Model}
From a computing perspective, each device $k$ requires the execution of a computational task characterized by an input data size $D_k^{\text{in}}$ and a computing intensity $\omega_k$ measured in CPU cycles per bit. The base station allocates $f_k^{\text{co}}$ CPU cycles per second to process the task of device $k$. The vector of computing resource allocations is denoted by $\mathbf{f}^\text{co} = [f^\text{co}_{1},\cdots,f^\text{co}_{K}]^T$, where the superscript `$\text{co}$' denotes computing-related variables. 

Due to the finite computing resources at the BS, the total computing capacity is limited by $F^{\text{max}}$. Let $p^{\text{co}}$ denote the total power consumed for computing all offloaded tasks. The computing-related constraints are collected in the set $\mathcal{D}_{\text{comp}}$, which includes the computing capacity constraint, the power–frequency relationship, and the resulting computing rate, which can be respectively written as:
\begin{align}
    \mathcal{D}_\text{comp}:\quad &\sideset{}{_{k\in\mathcal{K}}}\sum f^\text{co}_{k} \leq F^{\text{max}}, \label{eq:maxcompcap}\\
    &p^\text{co} = \tau \left(\sideset{}{_{k\in\mathcal{K}}}\sum f^\text{co}_{k} \right)^{\mu}, \label{eq:pco}\\
    &r_k^\text{co} = \frac{f^\text{co}_k}{\omega_k}, &&\forall k\in\mathcal{K}, \label{eq:rkco}
\end{align}
where $\tau$ and $\mu$ are hardware-dependent constants of the CPU power model \cite{10328648}.
The computing power is constrained by
\begin{align}
    \mathcal{P}_\text{comp}:\quad &p^\text{co} \leq P^\text{max,co},\label{eq:Pcomp}
\end{align}
where $P^{\text{max,co}}$ denotes the maximum available computing power. Finally, the computing delay is given by
\begin{align}
    \mathcal{T}_\text{comp}:\quad &t_k^\text{co} = \frac{D_k^\text{in}}{r_k^\text{co}}, &&\forall k\in\mathcal{K}.\label{eq:tcomp}
\end{align}
To best capture the joint impact of both communication and computing resources on the network performance, we next describe their dependency from rate, power and delay perspectives.

\subsubsection{Joint Communication and Computing}
In many emerging applications communication and computing processes are inherently coupled. For example, a head-mounted display may receive a 360-degree video stream that is pre-rendered and encoded at the base station \cite{10328648}. In such scenarios, network performance depends jointly on both communication and computing resources. 
Accordingly, the joint constraint sets introduced in the previous subsections can be combined as
\begin{align}
    \mathcal{D}_\text{joint}:\quad &\mathcal{D}_\text{comm} \cup \mathcal{D}_\text{comp},\\
    &r_k^\text{joint} = r_k^\text{tx} + r_k^\text{co}.
\end{align}

Moreover, the communication and computing power consumptions, namely the transmit power vector $\mathbf{p}^{\text{tx}}$ and the computing power $p^{\text{co}}$, are coupled at the BS. This coupling results in the following joint power constraint:
\begin{align}
    \mathcal{P}_\text{joint}:\quad &\sideset{}{_{k\in\mathcal{K}}}\sum p_k^\text{tx} + p^\text{co} \leq P^\text{max},\label{eq:P}
\end{align}
where $P^{\text{max}}$ denotes the maximum allowable power budget at the BS \cite{8330764}.

By jointly considering communication and computing delays, the total end-to-end latencies experienced by the devices are given by
\begin{align}
    \mathcal{T}_\text{joint}:\quad &t_k^\text{joint} = t_k^\text{tx}+t_k^\text{co}, &&\forall k\in\mathcal{K}.
\end{align}
Note that, the perfect knowledge of all network parameters is in general difficult to realize. Hence, we next adopt a practical perspective on parameter uncertainties.

\subsubsection{Robustness to Uncertainties}
We assume that the BS has access only to imperfect estimates of the wireless channels and computing requirements \cite{10071987}. Specifically, let $\hat{h}_{l,k}$ and $\hat{\omega}_k$ denote the estimated channel coefficient and computing intensity of device $k$, respectively. The true channel and computing parameters are modeled as
\begin{align}
    h_{l,k} &= \hat{h}_{l,k} + h^\text{err}_{l,k}, &&\forall(l,k)\in(\mathcal{L},\mathcal{K}),\label{eq:hlk}\\
    \omega_{k} &= \hat{\omega}_{k} + \omega^\text{err}_{k}, &&\forall k\in\mathcal{K},\label{eq:omegak}
\end{align}
where $h^{\text{err}}_{l,k} \sim \mathcal{CN}(0,\sigma_h^2)$ represents the channel estimation error and $\omega^{\text{err}}_k \sim \mathcal{N}(0,\sigma_\omega^2)$ denotes the computing requirement estimation error \cite{reifert2024robust}.

From both the communication and computing perspectives, robustness is incorporated through probabilistic performance constraints that explicitly account for the statistical distribution of the estimation errors and their impact on the network utility function $\lambda_{\text{utility}}(\mathbf{x})$. Here, $\lambda_{\text{utility}}(\mathbf{x})$ denotes the realized utility under channel and computation errors, while $f_{\text{utility}}(\mathbf{x})$ defines the optimization objective, which can be nominal or robust (e.g., based on the $\gamma$-th quantile). For communication-oriented optimization, the robust constraint set is defined as
\begin{align}
	\mathcal{R}_\text{comm}:\quad &\eqref{eq:hlk},\nonumber\\
    &\text{Pr}\left[\lambda_\text{utility}(\mathbf{x}) > \lambda_\text{utility}^\gamma(\mathbf{x}) \big| \mathbf{\hat{h}}_k, \forall k\in\mathcal{K} \right] \leq \gamma,\label{eq:percentile_comm}
\end{align}
where $\lambda_{\text{utility}}^{\gamma}(\mathbf{x})$ denotes the $\gamma$-th quantile of the utility distribution, i.e., the value for which only a fraction $\gamma$ of all uncertainty realizations yields a higher utility.

Similarly, for computing-oriented optimization, we define
\begin{align}
    \mathcal{R}_\text{comp}:\quad &\eqref{eq:omegak},\nonumber\\
    &\text{Pr}\left[\lambda_\text{utility}(\mathbf{x}) > \lambda_\text{utility}^\gamma(\mathbf{x}) \big| \hat{\omega}_k, \forall k\in\mathcal{K} \right] \leq \gamma.\label{eq:percentile_comp}
\end{align}

For joint communication and computing systems, the network utility depends on both the channel vectors $\mathbf{h}_k$ and the computing intensities $\omega_k$, each following potentially unknown probability distributions. Accordingly, the joint robust constraint set is expressed as
\begin{align}
    \mathcal{R}_\text{joint}:\quad &\eqref{eq:hlk},\eqref{eq:omegak},\nonumber\\
    &\text{Pr}\left[\lambda_\text{utility}(\mathbf{x}) > \lambda_\text{utility}^\gamma(\mathbf{x}) \big| \mathbf{\hat{h}}_k, \hat{\omega}_k, \forall k\in\mathcal{K} \right] \leq \gamma.\label{eq:percentile}
\end{align}

This percentile-based formulation ensures that the optimized solution satisfies the desired performance level with probability at least $1-\gamma$, thereby enabling robustness against statistical uncertainty while avoiding overly conservative worst-case designs. Based on the above models and concepts, we next describe the network optimization formulations.

\subsubsection{Network Optimization Formulation}
The proposed framework supports a broad class of utility optimization objectives within the generic formulation in \eqref{eq:Opt0}. These objectives are evaluated subject to a collection of constraints $\mathcal{C}(\mathbf{x})$, which may include a subset of the system model constraints introduced in the previous subsection.

In general, multiple network objectives can be captured by expert-specific utility functions $f_{\text{utility},a}(\mathbf{x}_a)$. In the context of the case study considered in this paper, we consider the following representative objectives: the sum rate $\sum_{k\in\mathcal{K}} r_k^{\nu}$, the minimum rate $\min_{k\in\mathcal{K}} (r_k^{\nu})$, the logarithmic rate $\sum_{k\in\mathcal{K}} \log(r_k^{\nu})$, the worst-case delay $\max_{k\in\mathcal{K}} (t_k^{\nu})$, and the sum delay $\sum_{k\in\mathcal{K}} t_k^{\nu}$. The latter two objectives are to be minimized, while the rate-based objectives are maximized. Here, $\nu \in \{\text{tx},\,\text{co},\,\text{joint}\}$ denotes whether the optimization targets communication, computing, or joint communication-computing performance, respectively.

Each objective can be formulated under either regular or robust optimization assumptions, depending on whether uncertainty in channel state information or computing requirements is taken into account. Accordingly, the expert-specific optimization variables $\mathbf{x}_n$ may include the transmit power allocation $\mathbf{p}^{\text{tx}}$, computing power $p^{\text{co}}$, CPU frequency allocation $\mathbf{f}^{\text{co}}$, or appropriate subsets thereof.

Table~\ref{tab:experts} presents the complete set of $30$ experts employed in the case study considered in this paper. Furthermore, Fig.~\ref{fig:cake} (in the appendix) provides a qualitative illustration of the distinct characteristics of each expert in terms of robustness, communication and computing focus, as well as their emphasis on rate, delay, throughput, and fairness objectives.

This diverse expert library, i.e., Table~\ref{tab:experts}, enables the LLM-based gate to reason over heterogeneous optimization strategies and construct adaptive mixtures tailored to varying network objectives and uncertainty levels. Next, we describe how such network optimization problems, i.e., those related to Agents $1$-$30$ in Table~\ref{tab:experts}, are solved using deep learning-based agents.

\begin{table*}[tb]
    \centering
    \caption{List of available experts with index $a\in\mathcal{N}$, name, role, and system impact, as well as the corresponding optimization formulation in terms of variables ($\mathbf{x}_a$), objective function ($f_{\text{utility},a}(\mathbf{x}_a)$), constraint set ($\mathcal{C}_a(\mathbf{x}_a)$), and robustness utility ($\lambda_{\text{utility},a}$).}
    \label{tab:experts}
    \resizebox{\linewidth}{!}{\begin{tabular}{c l l l l l l l}
        \cellcolor{black!50}\textcolor{white}{\textbf{}{Idx.}} & 
        \cellcolor{black!50}\textcolor{white}{\textbf{Name}} & 
        \cellcolor{black!50}\textcolor{white}{\textbf{Role}} & 
        \cellcolor{black!50}\textcolor{white}{\textbf{System Impact}}& 
        \cellcolor{black!50}\textcolor{white}{\textbf{$\mathbf{x}_a$}}&
        \cellcolor{black!50}\textcolor{white}{\textbf{$f_{\text{utility},a}(\mathbf{x}_a)$}}&
        \cellcolor{black!50}\textcolor{white}{\textbf{$\mathcal{C}_a(\mathbf{x}_a)$}}&
        \cellcolor{black!50}\textcolor{white}{\textbf{$\lambda_{\text{utility},a}$}}\\ 
        $1$ & 
        Comm\_SumR\_Reg & 
        Comm. sum rate max. & 
        Throughput &
        $\mathbf{p}^\text{tx}$ &
        $\sum_{k\in\mathcal{K}}r_k^\text{tx}$ &
        $\mathcal{D}_\text{comm} \cup \mathcal{P}_\text{comm}$ &
        --\\
        \cellcolor{black!10}$2$ & 
        \cellcolor{black!10}Comm\_SumR\_Rob & 
        \cellcolor{black!10}Comm. sum rate max. & 
        \cellcolor{black!10}Throughput & 
        \cellcolor{black!10}$\mathbf{p}^\text{tx}$ &
        \cellcolor{black!10}$\lambda_{\text{sumR},a}^\gamma$ & 
        \cellcolor{black!10}$\mathcal{D}_\text{comm} \cup \mathcal{P}_\text{comm} \cup \mathcal{R}_\text{comm}$ & 
        \cellcolor{black!10}$\sum_{k\in\mathcal{K}}r_k^\text{tx}$\\ 
        $3$ & 
        Comp\_SumR\_Reg & 
        Comp. sum rate max. & 
        Throughput &
        $p^\text{co}, \mathbf{f}^\text{co}$ &
        $\sum_{k\in\mathcal{K}}r_k^\text{co}$ &
        $\mathcal{D}_\text{comp} \cup \mathcal{P}_\text{comp}$ &
        --\\
        \cellcolor{black!10}$4$ & 
        \cellcolor{black!10}Comp\_SumR\_Rob & 
        \cellcolor{black!10}Comp. sum rate max. & 
        \cellcolor{black!10}Throughput & 
        \cellcolor{black!10}$p^\text{co}, \mathbf{f}^\text{co}$ &
        \cellcolor{black!10}$\lambda_{\text{sumR},a}^\gamma$ & 
        \cellcolor{black!10}$\mathcal{D}_\text{comp} \cup \mathcal{P}_\text{comp} \cup \mathcal{R}_\text{comp}$ & 
        \cellcolor{black!10}$\sum_{k\in\mathcal{K}}r_k^\text{co}$\\ 
        $5$ & 
        JCC\_SumR\_Reg & 
        Sum rate max. & 
        Throughput &
        $\mathbf{p}^\text{tx}, p^\text{co}, \mathbf{f}^\text{co}$ &
        $\sum_{k\in\mathcal{K}}(r_k^\text{tx} + r_k^\text{co})$ &
        $\mathcal{D}_\text{comp} \cup \mathcal{P}_\text{comp}$ &
        --\\
        \cellcolor{black!10}$6$ & 
        \cellcolor{black!10}JCC\_SumR\_Rob & 
        \cellcolor{black!10}Sum rate max. & 
        \cellcolor{black!10}Throughput & 
        \cellcolor{black!10}$\mathbf{p}^\text{tx}, p^\text{co}, \mathbf{f}^\text{co}$ &
        \cellcolor{black!10}$\lambda_{\text{sumR},a}^\gamma$ & 
        \cellcolor{black!10}$\mathcal{D}_\text{joint} \cup \mathcal{P}_\text{joint} \cup \mathcal{R}_\text{joint}$ & 
        \cellcolor{black!10}$\sum_{k\in\mathcal{K}}(r_k^\text{tx} + r_k^\text{co})$\\ 
        $7$ & 
        Comm\_MinR\_Reg & 
        Comm. min rate max. & 
        Fairness &
        $\mathbf{p}^\text{tx}$ &
        $\min_{k\in\mathcal{K}}(r_k^\text{tx})$ &
        $\mathcal{D}_\text{comm} \cup \mathcal{P}_\text{comm}$ &
        --\\
        \cellcolor{black!10}$8$ & 
        \cellcolor{black!10}Comm\_MinR\_Rob & 
        \cellcolor{black!10}Comm. min rate max. & 
        \cellcolor{black!10}Fairness & 
        \cellcolor{black!10}$\mathbf{p}^\text{tx}$ &
        \cellcolor{black!10}$\lambda_{\text{minR},a}^\gamma$ & 
        \cellcolor{black!10}$\mathcal{D}_\text{comm} \cup \mathcal{P}_\text{comm} \cup \mathcal{R}_\text{comm}$ & 
        \cellcolor{black!10}$\min_{k\in\mathcal{K}}(r_k^\text{tx})$\\ 
        $9$ & 
        Comp\_MinR\_Reg & 
        Comp. min rate max. & 
        Fairness &
        $p^\text{co}, \mathbf{f}^\text{co}$ &
        $\min_{k\in\mathcal{K}}(r_k^\text{co})$ &
        $\mathcal{D}_\text{comp} \cup \mathcal{P}_\text{comp}$ &
        --\\
        \cellcolor{black!10}$10$ & 
        \cellcolor{black!10}Comp\_MinR\_Rob & 
        \cellcolor{black!10}Comp. min rate max. & 
        \cellcolor{black!10}Fairness & 
        \cellcolor{black!10}$p^\text{co}, \mathbf{f}^\text{co}$ &
        \cellcolor{black!10}$\lambda_{\text{minR},a}^\gamma$ & 
        \cellcolor{black!10}$\mathcal{D}_\text{comp} \cup \mathcal{P}_\text{comp} \cup \mathcal{R}_\text{comp}$ & 
        \cellcolor{black!10}$\min_{k\in\mathcal{K}}(r_k^\text{co})$\\ 
        $11$ & 
        JCC\_MinR\_Reg & 
        Min rate max. & 
        Fairness &
        $\mathbf{p}^\text{tx}, p^\text{co}, \mathbf{f}^\text{co}$ &
        $\min_{k\in\mathcal{K}}(r_k^\text{tx} + r_k^\text{co})$ &
        $\mathcal{D}_\text{comp} \cup \mathcal{P}_\text{comp}$ &
        --\\
        \cellcolor{black!10}$12$ & 
        \cellcolor{black!10}JCC\_MinR\_Rob & 
        \cellcolor{black!10}Min rate max. & 
        \cellcolor{black!10}Fairness & 
        \cellcolor{black!10}$\mathbf{p}^\text{tx}, p^\text{co}, \mathbf{f}^\text{co}$ &
        \cellcolor{black!10}$\lambda_{\text{minR},a}^\gamma$ & 
        \cellcolor{black!10}$\mathcal{D}_\text{joint} \cup \mathcal{P}_\text{joint} \cup \mathcal{R}_\text{joint}$ & 
        \cellcolor{black!10}$\min_{k\in\mathcal{K}}(r_k^\text{tx} + r_k^\text{co})$\\ 
        $13$ & 
        Comm\_LogR\_Reg & 
        Comm. log rate max. & 
        Balanced &
        $\mathbf{p}^\text{tx}$ &
        $\sum_{k\in\mathcal{K}}\log(r_k^\text{tx})$ &
        $\mathcal{D}_\text{comm} \cup \mathcal{P}_\text{comm}$ &
        --\\
        \cellcolor{black!10}$14$ & 
        \cellcolor{black!10}Comm\_LogR\_Rob & 
        \cellcolor{black!10}Comm. log rate max. & 
        \cellcolor{black!10}Balanced & 
        \cellcolor{black!10}$\mathbf{p}^\text{tx}$ &
        \cellcolor{black!10}$\lambda_{\text{logR},a}^\gamma$ & 
        \cellcolor{black!10}$\mathcal{D}_\text{comm} \cup \mathcal{P}_\text{comm} \cup \mathcal{R}_\text{comm}$ & 
        \cellcolor{black!10}$\sum_{k\in\mathcal{K}}\log(r_k^\text{tx})$\\ 
        $15$ & 
        Comp\_LogR\_Reg & 
        Comp. log rate max. & 
        Balanced &
        $p^\text{co}, \mathbf{f}^\text{co}$ &
        $\sum_{k\in\mathcal{K}}\log(r_k^\text{co})$ &
        $\mathcal{D}_\text{comp} \cup \mathcal{P}_\text{comp}$ &
        --\\
        \cellcolor{black!10}$16$ & 
        \cellcolor{black!10}Comp\_LogR\_Rob & 
        \cellcolor{black!10}Comp. log rate max. & 
        \cellcolor{black!10}Balanced & 
        \cellcolor{black!10}$p^\text{co}, \mathbf{f}^\text{co}$ &
        \cellcolor{black!10}$\lambda_{\text{logR},a}^\gamma$ & 
        \cellcolor{black!10}$\mathcal{D}_\text{comp} \cup \mathcal{P}_\text{comp} \cup \mathcal{R}_\text{comp}$ & 
        \cellcolor{black!10}$\sum_{k\in\mathcal{K}}\log(r_k^\text{co})$\\ 
        $17$ & 
        JCC\_LogR\_Reg & 
        Log rate max. & 
        Balanced &
        $\mathbf{p}^\text{tx}, p^\text{co}, \mathbf{f}^\text{co}$ &
        $\sum_{k\in\mathcal{K}}\log(r_k^\text{tx} + r_k^\text{co})$ &
        $\mathcal{D}_\text{comp} \cup \mathcal{P}_\text{comp}$ &
        --\\
        \cellcolor{black!10}$18$ & 
        \cellcolor{black!10}JCC\_LogR\_Rob & 
        \cellcolor{black!10}Log rate max. & 
        \cellcolor{black!10}Balanced & 
        \cellcolor{black!10}$\mathbf{p}^\text{tx}, p^\text{co}, \mathbf{f}^\text{co}$ &
        \cellcolor{black!10}$\lambda_{\text{logR},a}^\gamma$ & 
        \cellcolor{black!10}$\mathcal{D}_\text{joint} \cup \mathcal{P}_\text{joint} \cup \mathcal{R}_\text{joint}$ & 
        \cellcolor{black!10}$\sum_{k\in\mathcal{K}}\log(r_k^\text{tx} + r_k^\text{co})$\\
        $19$ & 
        Comm\_MaxT\_Reg & 
        Comm. max delay min. & 
        Responsiveness &
        $\mathbf{p}^\text{tx}$ &
        $\sum_{k\in\mathcal{K}}(t_k^\text{tx})$ &
        $\mathcal{D}_\text{comm} \cup \mathcal{P}_\text{comm} \cup \mathcal{T}_\text{comm}$ &
        --\\
        \cellcolor{black!10}$20$ & 
        \cellcolor{black!10}Comm\_MaxT\_Rob & 
        \cellcolor{black!10}Comm. max delay min. & 
        \cellcolor{black!10}Responsiveness & 
        \cellcolor{black!10}$\mathbf{p}^\text{tx}$ &
        \cellcolor{black!10}$\lambda_{\text{maxT},a}^\gamma$ & 
        \cellcolor{black!10}$\mathcal{D}_\text{comm} \cup \mathcal{P}_\text{comm} \cup \mathcal{R}_\text{comm} \cup \mathcal{T}_\text{comm}$ & 
        \cellcolor{black!10}$\max_{k\in\mathcal{K}}(t_k^\text{tx})$\\ 
        $21$ & 
        Comp\_MaxT\_Reg & 
        Comp. max delay min. & 
        Responsiveness &
        $p^\text{co}, \mathbf{f}^\text{co}$ &
        $\max_{k\in\mathcal{K}}(t_k^\text{co})$ &
        $\mathcal{D}_\text{comp} \cup \mathcal{P}_\text{comp} \cup \mathcal{T}_\text{comp}$ &
        --\\
        \cellcolor{black!10}$22$ & 
        \cellcolor{black!10}Comp\_MaxT\_Rob & 
        \cellcolor{black!10}Comp. max delay min. & 
        \cellcolor{black!10}Responsiveness & 
        \cellcolor{black!10}$p^\text{co}, \mathbf{f}^\text{co}$ &
        \cellcolor{black!10}$\lambda_{\text{maxT},a}^\gamma$ & 
        \cellcolor{black!10}$\mathcal{D}_\text{comp} \cup \mathcal{P}_\text{comp} \cup \mathcal{R}_\text{comp} \cup \mathcal{T}_\text{comp}$ & 
        \cellcolor{black!10}$\max_{k\in\mathcal{K}}(t_k^\text{co})$\\ 
        $23$ & 
        JCC\_MaxT\_Reg & 
        Max delay min. & 
        Responsiveness &
        $\mathbf{p}^\text{tx}, p^\text{co}, \mathbf{f}^\text{co}$ &
        $\max_{k\in\mathcal{K}}(t_k^\text{tx} + t_k^\text{co})$ &
        $\mathcal{D}_\text{comp} \cup \mathcal{P}_\text{comp} \cup \mathcal{T}_\text{joint}$ &
        --\\
        \cellcolor{black!10}$24$ & 
        \cellcolor{black!10}JCC\_MaxT\_Rob & 
        \cellcolor{black!10}Max delay min. & 
        \cellcolor{black!10}Responsiveness & 
        \cellcolor{black!10}$\mathbf{p}^\text{tx}, p^\text{co}, \mathbf{f}^\text{co}$ &
        \cellcolor{black!10}$\lambda_{\text{maxT},a}^\gamma$ & 
        \cellcolor{black!10}$\mathcal{D}_\text{joint} \cup \mathcal{P}_\text{joint} \cup \mathcal{R}_\text{joint} \cup \mathcal{T}_\text{joint}$ & 
        \cellcolor{black!10}$\max_{k\in\mathcal{K}}(t_k^\text{tx} + t_k^\text{co})$\\
        $25$ & 
        Comm\_SumT\_Reg & 
        Comm. sum delay min. & 
        Latency &
        $\mathbf{p}^\text{tx}$ &
        $\sum_{k\in\mathcal{K}}(t_k^\text{tx})$ &
        $\mathcal{D}_\text{comm} \cup \mathcal{P}_\text{comm} \cup \mathcal{T}_\text{comm}$ &
        --\\
        \cellcolor{black!10}$26$ & 
        \cellcolor{black!10}Comm\_SumT\_Rob & 
        \cellcolor{black!10}Comm. sum delay min. & 
        \cellcolor{black!10}Latency & 
        \cellcolor{black!10}$\mathbf{p}^\text{tx}$ &
        \cellcolor{black!10}$\lambda_{\text{sumT},a}^\gamma$ & 
        \cellcolor{black!10}$\mathcal{D}_\text{comm} \cup \mathcal{P}_\text{comm} \cup \mathcal{R}_\text{comm} \cup \mathcal{T}_\text{comm}$ & 
        \cellcolor{black!10}$\sum_{k\in\mathcal{K}}(t_k^\text{tx})$\\ 
        $27$ & 
        Comp\_SumT\_Reg & 
        Comp. sum delay min. & 
        Latency &
        $p^\text{co}, \mathbf{f}^\text{co}$ &
        $\sum_{k\in\mathcal{K}}(t_k^\text{co})$ &
        $\mathcal{D}_\text{comp} \cup \mathcal{P}_\text{comp} \cup \mathcal{T}_\text{comp}$ &
        --\\
        \cellcolor{black!10}$28$ & 
        \cellcolor{black!10}Comp\_SumT\_Rob & 
        \cellcolor{black!10}Comp. sum delay min. & 
        \cellcolor{black!10}Latency & 
        \cellcolor{black!10}$p^\text{co}, \mathbf{f}^\text{co}$ &
        \cellcolor{black!10}$\lambda_{\text{sumT},a}^\gamma$ & 
        \cellcolor{black!10}$\mathcal{D}_\text{comp} \cup \mathcal{P}_\text{comp} \cup \mathcal{R}_\text{comp} \cup \mathcal{T}_\text{comp}$ & 
        \cellcolor{black!10}$\sum_{k\in\mathcal{K}}(t_k^\text{co})$\\ 
        $29$ & 
        JCC\_SumT\_Reg & 
        Sum delay min. & 
        Latency &
        $\mathbf{p}^\text{tx}, p^\text{co}, \mathbf{f}^\text{co}$ &
        $\sum_{k\in\mathcal{K}}(t_k^\text{tx} + t_k^\text{co})$ &
        $\mathcal{D}_\text{comp} \cup \mathcal{P}_\text{comp} \cup \mathcal{T}_\text{joint}$ &
        --\\
        \cellcolor{black!10}$30$ & 
        \cellcolor{black!10}JCC\_SumT\_Rob & 
        \cellcolor{black!10}Sum delay min. & 
        \cellcolor{black!10}Latency & 
        \cellcolor{black!10}$\mathbf{p}^\text{tx}, p^\text{co}, \mathbf{f}^\text{co}$ &
        \cellcolor{black!10}$\lambda_{\text{sumT},a}^\gamma$ & 
        \cellcolor{black!10}$\mathcal{D}_\text{joint} \cup \mathcal{P}_\text{joint} \cup \mathcal{R}_\text{joint} \cup \mathcal{T}_\text{joint}$ & 
        \cellcolor{black!10}$\sum_{k\in\mathcal{K}}(t_k^\text{tx} + t_k^\text{co})$
    \end{tabular}}
\end{table*}

\subsection{Deep Learning-based Experts}\label{ssec:B}
To enable real-time inference, the optimization experts introduced in the previous subsection, i.e., Agents $1$-$30$, are implemented using deep neural network (DNN) agents. For the sake of conciseness, rather than defining separate network architectures for each expert, we adopt a unified expert template in which each expert corresponds to a parameterized neural optimization operator.

\begin{figure}[t]
\centering
\includegraphics[width=\linewidth]{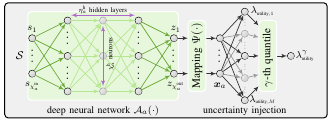}\vspace*{-.2cm}
\caption{Generic DNN structure and uncertainty injection mechanism utilized in the proposed case study.}
\label{fig:dnnmdl}
\vspace*{-.4cm}
\end{figure}

Specifically, expert $a$ is modeled as
\begin{align}
    \mathbf{z}_a = \mathcal{A}_a(\mathcal{S}; \boldsymbol{\Theta}_a),
\end{align}
where $\mathcal{S}$ denotes the network state, $\mathcal{A}_a(\cdot)$ represents a generic DNN architecture, and $\boldsymbol{\Theta}_a$ denotes the parameter set associated with expert $a$, as illustrated in Fig.~\ref{fig:dnnmdl}.

Each expert DNN contains a total of $|\boldsymbol{\Theta}_a|$ trainable parameters. The network architecture consists of $\eta_a^{\text{h}}$ fully connected hidden layers, each with $\chi_a^{\text{h}}$ neurons, $\chi_a^{\text{in}}$ input neurons, and $\chi_a^{\text{out}}$ output neurons. The total number of trainable parameters, including weights and biases, is given by
\begin{align}
|\boldsymbol{\Theta}_a| &= \chi_a^{\text{in}} \chi_a^{\text{h}} + (\eta_a^{\text{h}} - 1) (\chi_a^{\text{h}})^2 + \chi_a^{\text{out}} \chi_a^{\text{h}} \nonumber \\
&\qquad + \eta_a^{\text{h}} \chi_a^{\text{h}} + \chi_a^{\text{out}}.
\end{align}

The DNN input is given by the vector $\mathbf{s} = [s_1,\ldots,s_{\chi_a^{\text{in}}}]^T$, dependent on the network state $\mathcal{S}.$ The input layer dimension $\chi_a^{\text{in}}$ depends on the expert specialization and satisfies
\[
\chi_a^{\text{in}} \in \{2K^2,\; K,\; 2K^2 + K\},
\]
corresponding to communication-focused, computing-focused, and joint communication and computing experts, respectively. For communication-focused experts, the input consists of the real and imaginary components of the effective channel coefficients, resulting in $2K^2$ input features. For computing-focused experts, the input consists of the $K$ task-related parameters (e.g., computing intensities or workload descriptors). For joint communication and computing experts, both sets of features are included, yielding $2K^2 + K$ input features that together represent the network state $\mathcal{S}$.

The input layer is followed by $\eta_a^{\text{h}}$ fully connected hidden layers with rectified linear unit (ReLU) activation functions.

The output layer dimension depends on the expert type and satisfies
\[
\chi_a^{\text{out}} \in \{K,\; K+1,\; 2K+1\}.
\]
For communication-oriented experts, the $K$ output neurons correspond to the transmit powers allocated to the users. For computing-oriented experts, $K$ neurons represent the allocated computing capacities and one neuron represents the total computing power. For joint experts, the first $K+1$ neurons correspond to the transmit powers allocated to all users and the total computing power, while the subsequent $K$ neurons represent the allocated computing capacities.

The output vector $\mathbf{z} = [z_1,\ldots,z_{\chi_a^{\text{out}}}]^T$ is processed by a dedicated mapping layer that transforms the neural outputs into feasible optimization variables $(\mathbf{p}^{\text{tx}}, p^{\text{co}}, \mathbf{f}^{\text{co}})$, or appropriate subsets thereof, while ensuring compliance with system constraints as
\begin{equation}
    \mathbf{x}_a = \Psi(\mathbf{z}_a),
\end{equation}
where $\Psi$ is the mapping layer.

\subsubsection{Mapping Layer}
For completeness, we describe a representative implementation of the mapping layer $\Psi(\cdot)$ used in the case study to transform neural outputs into feasible resource allocations. The transmit powers and total computing power are obtained by applying a softmax mapping\footnote{For an input vector $\mathbf{x}$, the softmax function outputs a vector whose $i$-th element is $e^{x_i}/\sum_{j=1}^{n} e^{x_j}$, where $n$ is the input dimension.} to the corresponding DNN outputs
\begin{equation}
	[(\mathbf{p}^\text{tx})^T,p^\text{co}]^T = P^\text{max}\cdot \text{softmax}\big([z_{1},\ldots,z_{K+1}]^T\big),\label{eq:ptx_pco}
\end{equation}
which distributes the available power budget across communication and computing resources. The computing power $p^{\text{co}}$ is then translated into the maximum available computing cycles using the CPU power model in \eqref{eq:pco}
\begin{equation}
	F^\text{pow} = \sqrt[\leftroot{1}\uproot{1}\mu]{\frac{p^\text{co}}{\tau}}.
\end{equation}
Since the effective computing allocation is limited by both the platform capacity $F^{\text{max}}$ and the power-dependent limit $F^{\text{pow}}$, the final computing allocation vector is obtained as
\begin{equation}
	\mathbf{f}^\text{co} = \text{min}\big(F^\text{max},F^\text{pow}\big)\cdot \text{softmax}\big([z_{K+2},\ldots,z_{2K+1}]^T\big),\label{eq_8:fco}
\end{equation}
thereby enforcing feasibility while proportionally distributing the available computing cycles among users. Note that the mapping operations may vary across experts. For instance, communication-focused experts may only apply \eqref{eq:ptx_pco} without including the computing power variable $p^{\text{co}}$. More generally, alternative mapping functions can be incorporated without modifying the proposed agentic MoE framework.

\subsubsection{Uncertainty Injection}
For robust experts, i.e., those which account for channel or computing uncertainty, we employ an uncertainty injection mechanism to account for stochastic channel and computing estimation errors \cite{10071987,reifert2024robust}. An illustration of the uncertainty injection mechanism can be found on the right hand side within Fig.~\ref{fig:dnnmdl}. The uncertainty injection principle approximates statistical performance objectives numerically by evaluating them over multiple uncertainty realizations. Specifically, uncertainty samples are injected after the DNN output, enabling the evaluation of multiple realizations of the resulting optimization variables. From these realizations, an empirical robust objective is constructed. Channel estimation errors are modeled as $\mathbf{h}^{\text{err}}_{k} = [h^{\text{err}}_{1,k}, \ldots, h^{\text{err}}_{L,k}]^T$, while computing estimation errors are denoted by $\omega^{\text{err}}_k$. Both uncertainty sources are assumed to be sampleable. Accordingly, $M$ independent realizations $\{\mathbf{h}^{\text{err}}_{k,m}, \omega^{\text{err}}_{k,m}\}_{m=1}^{M}$ are generated for all $k \in \mathcal{K}$. For each realization $m$, the resulting performance utility $\lambda_{\text{utility},m}$ is evaluated. The robust utility is then approximated using the empirical $\gamma$-th quantile
\begin{align}
    \lambda_{\text{utility}}^{\gamma} = \text{percentile}_{\gamma} \left(\lambda_{\text{utility},1}, \ldots, \lambda_{\text{utility},M}\right),
\end{align}
which estimates the $\gamma$-th quantile performance under uncertainty.

\subsubsection{Training Procedure}
The expert parameters $\boldsymbol{\Theta}_a$ are obtained by training the DNN to approximate solutions of the corresponding optimization problem under specific objectives, constraint configurations, and uncertainty models. This is achieved by minimizing an expert-specific loss function
\begin{align}
    \boldsymbol{\Theta}_a = \arg\min_{\boldsymbol{\Theta}_a} \mathbb{E}_{\mathcal{S}} \left[\mathcal{L}_a\big(\mathcal{A}_a(\mathcal{S};\boldsymbol{\Theta_a})\big)\right],
\end{align}
where $\mathcal{L}_a(\cdot)$ encodes the optimization objective and performance criteria associated with expert $a$. 

This formulation allows the expert library to capture diverse optimization strategies, that are fundamental to 6G systems operations, namely, throughput-oriented, fairness-aware, delay-driven, and robustness-focused resource allocation policies, while maintaining a unified neural architecture. Different experts are, therefore, distinguished primarily through their training objectives and data distributions rather than architectural differences, which facilitates scalable expert composition within the agentic AI-based framework.

\subsubsection{LLM-enabled Gate Network}
We now describe the case study-specific realization of the LLM-enabled gate network introduced in Section~\ref{sec:2}. In this work, the LLM acts as an agentic decision-making layer that interprets high-level operator objectives and selects appropriate optimization experts from the available expert library.

The system prompt $\mathcal{Q}^{\text{sys}}$ defines the available optimization experts, their respective capabilities, and the expected output format of the gate. In particular, the prompt describes each expert in terms of its optimization objective, robustness properties, and applicable operating conditions. For the detailed system prompt used within this case study, we refer to the appendix of this paper.

Given the system prompt and the operator query $\mathcal{Q}^{\text{op}}(\mathcal{S})$, the LLM generates a structured tool call that specifies the selected expert(s) and their corresponding weights via the selection vector $\mathbf{a}$ and weighting vector $\boldsymbol{\alpha}$ according to \eqref{eq:LLM}, as described in Section~\ref{sec:2} above. The resulting expert selections and weights are used to generate the final resource allocation decisions according to equation \eqref{eq:combexp}. 

For illustration purposes, we next describe the problem formulation and solution approach for one representative agent.

\subsection{Example Problem and Solution: Agent $24$}\label{ssec:C}
By way of example, we now present the details of one particular agent, namely Agent~$24$ as described in Table~\ref{tab:experts}, for showcasing the steps of the expert design and training procedure. Specifically, we consider a joint communication and computing worst-case delay minimization problem under channel and computing uncertainty conditions. The corresponding optimization problem is given by
\begin{subequations}\label{eq:OptExample}
\begin{align}
	\underset{\mathbf{p}^\text{tx}, p^\text{co}, \mathbf{f}^\text{co}}{\text{max}}\quad &\lambda_{\text{maxT}}^\gamma\tag{\ref{eq:OptExample}} \\
    \text{s.t.} \quad\quad & \eqref{eq:hlk},\eqref{eq:omegak},\mathcal{D}_\text{joint},\mathcal{P}_\text{joint},\mathcal{T}_\text{joint},\nonumber\\
    &\text{Pr}\left[\max_{k\in\mathcal{K}}(t_k^\text{joint}) > \lambda_{\text{maxT}}^\gamma\big| \mathbf{\hat{h}}_k, \hat{\omega}_k, \forall k\in\mathcal{K} \right] \leq \gamma.\nonumber\\[-.3cm]\label{eq:24}
\end{align}
\end{subequations}
Here, \eqref{eq:hlk} and \eqref{eq:omegak} model the channel and computing estimation errors, respectively, while $\mathcal{D}_\text{joint}$, $\mathcal{P}_\text{joint}$, and $\mathcal{T}_\text{joint}$ capture the communication and computing rate, power, and delay constraints. Constraint~\eqref{eq:24} enforces a $\gamma$-th quantile (robust) constraint on the worst-case joint communication and computing delay.

Problem~\eqref{eq:OptExample} is addressed using a DNN-based approximation. The corresponding agent is denoted by $\mathcal{A}_{24}(\mathcal{S}; \boldsymbol{\Theta}_{24})$. The input dimension is $\chi_n^{\text{in}} = 2K^2 + K$, comprising the real and imaginary parts of the effective channel gains $\{\mathbf{\hat{h}}_{k,j}^\text{eff} = \mathbf{\hat{h}}_k^H \mathbf{v}_j\}_{k,j\in\mathcal{K}}$ and the computing intensities $\{\hat{\omega}_k\}_{k\in\mathcal{K}}$. The network output is $\mathbf{z} \in \mathbb{R}^{(2K+1)\times 1}$ and is mapped to the optimization variables $\mathbf{x}_{24} = [(\mathbf{p}^{\text{tx}})^T,\, p^{\text{co}},\, (\mathbf{f}^{\text{co}})^T]^T$ via \eqref{eq:ptx_pco} and \eqref{eq_8:fco}. To account for uncertainty, an uncertainty injection layer is applied, and the loss function $\mathcal{L}_{24}(\cdot)$ corresponds to the robust objective $\lambda_{\text{maxT}}^\gamma$. Further details on the training procedure can be found in \cite{reifert2024robust}. We next evaluate the case study through numerical simulations.

\section{Case Study: Simulation Results}\label{sec:cs2}
\begin{table}[t]
\centering
\caption{Simulation parameters.}\label{tb:simparam}\vspace*{-.2cm}
\resizebox{\linewidth}{!}{\begin{tabular}{l l}
	\cellcolor{black!50}\textcolor{white}{\textbf{Network resources}} & \cellcolor{black!50}\textcolor{white}{\textbf{DNN parameters}}\\
	\cellcolor{black!10}$L=4$ antennas, $K=4$ users & \cellcolor{black!10}$\eta_n^h = 10$ layers (ReLU)\\
	$P^\text{max}=34$ dBm & $\chi_n^h = 400$ neurons per layer\\
	\cellcolor{black!10}$W=5$ MHz, $\Gamma^\text{gap}=9.5$ dB &  \cellcolor{black!10}$500$ epochs, $50$ minibatches\\
	$\sigma^2 = -75$ dBm/Hz & $1000$ batch \& $2000$ validation size\\
	\cellcolor{black!10}$F^\text{max}=4.6$ GHz, $\tau=10^{-28}$, $\mu=3$ & \cellcolor{black!10}$4000$ testing size\\
    $D_k^\text{out}=2.5\cdot 10^4$ bit, $D_k^\text{in}=5\cdot 10^4$ bit & \\
    \multicolumn{2}{l}{\cellcolor{black!10}$\sigma_h^2=0.15$, $\sigma_\omega^2 = 3200$, $\gamma=0.05$, $\alpha=0.2$ regularized zero-forcing factor}
\end{tabular}}
\end{table}%
This section provides numerical evaluations of the proposed agentic AI-based optimization framework in the context of the illustrated joint computing and communication systems optimization presented in Section~\ref{sec:cs}. The simulations particularly highlight how the LLM-enabled MoE architecture selects and combines specialized optimization agents to address diverse network objectives and uncertainty conditions. Multiple simulation sets are considered to evaluate optimization performance, the quality of expert compositions, and the semantic accuracy of the LLM-based agent selection. The corresponding system parameters for each scenario are summarized in the respective tables.

The simulation parameters are summarized in Table~\ref{tb:simparam}. We assume Rayleigh fading $h_{l,k}\sim \mathcal{CN}(0,1)$, $\forall (l,k)\in(\mathcal{L},\mathcal{K})$, utilize minimum mean-square estimation for the estimated channels and set the beamforming vectors according to regularized zero-forcing \cite{10071987}. The ground-truth computing intensities are drawn according to a Gamma distribution, i.e., $\omega_{k}\sim \Gamma(2,200)$, $\forall k\in\mathcal{K}$. We note that details of the adopted training procedure can be found in \cite{asilomar_EE, reifert2024robust}.

\subsubsection{Simulation Set 1}
\begin{table}[t]
    \centering
    \caption{Description of simulation Set 1.}\label{tab:simset1}\vspace*{-.2cm}
    \resizebox{\linewidth}{!}{\begin{tabular}{r p{6.5cm}}
        \multicolumn{2}{l}{\cellcolor{black!50}\textcolor{white}{\textbf{Framework setup}}}\\
        \cellcolor{black!50}\textcolor{white}{\textbf{Query}} & \cellcolor{black!10}\emph{I have connected extended reality devices. I am rendering their field of view and then transmitting the video stream. All users should be treated fair and have minimal latency}\\
        \cellcolor{black!50}\textcolor{white}{\textbf{Expert library}} & Experts $1$-$4$, $7$-$10$, $13$-$16$, and $19$-$22$\\
        \cellcolor{black!50}\textcolor{white}{\textbf{Optimal benchmarks}} & \cellcolor{black!10}Experts $5$, $6$, $11$, $12$, $17$, $18$, $23$, and $24$\\
        \multicolumn{2}{l}{\cellcolor{black!50}\textcolor{white}{\textbf{Agentic AI solution}}}\\
        \cellcolor{black!50}\textcolor{white}{\textbf{Interpretation}} & \cellcolor{black!10}\emph{To address your query, we need to focus on minimizing the maximum delay for both communication and computing, ensuring fairness among all users. Since you are dealing with both rendering (computing) and video transmission (communication), we should consider a joint approach. Given that you want minimal latency and fairness, the experts for minimizing maximum delay under regular conditions for both communication and computing should be used.}\\
        \cellcolor{black!50}\textcolor{white}{\textbf{Selection}} & $19$ \emph{Comm\_MaxT\_Reg} \& $21$ \emph{Comp\_MaxT\_Reg}\\
        \cellcolor{black!50}\textcolor{white}{\textbf{Weights}} & \cellcolor{black!10}$\alpha_{19} = \alpha_{21} = 0.5$
    \end{tabular}}
\end{table}
The setup of simulation Set~1 is given in Table~\ref{tab:simset1}. The optimizer library, i.e., a subset of Table~\ref{tab:experts}, consists of sum-rate, min-rate, log-rate, and worst-case delay experts trained separately for communication and computing objectives. Joint communication and computing experts serve as optimal benchmarking schemes, which allows to evaluate how well the proposed agentic AI framework can approximate joint communication and computing optimization by composing separate experts from the library.

\begin{figure}[t]
\centering
\includegraphics[width=.9\linewidth]{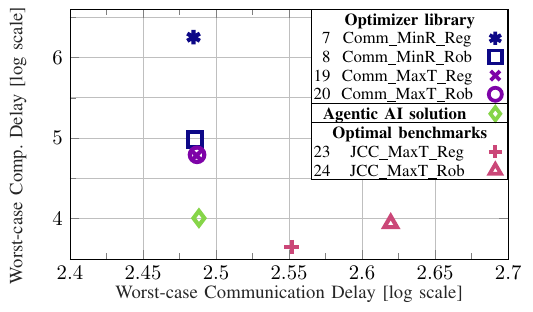}\vspace*{-.2cm}
\caption{Experts, benchmarks, and agentic AI solution space as scatter plot, including infeasible results for simulation Set 1.}
\label{fig:set1_scatter_single}
\vspace*{-.4cm}
\end{figure}
Fig.~\ref{fig:set1_scatter_single} illustrates the feasible solution space in terms of worst-case computing delay $\log(\max_{k\in\mathcal{K}}(t_k^\text{co}))$ and worst-case communication delay $\log(\max_{k\in\mathcal{K}}(t_k^\text{tx}))$. The figure includes individual expert solutions, benchmark solutions, and the proposed agentic AI solution. Only solutions within the feasible delay region are shown. The results demonstrate that the proposed agentic AI framework outperforms all individual library experts in both communication and computing delay. Furthermore, the agentic AI solution improves upon conventional benchmark schemes such as sum-rate and min-rate optimization, which may yield infeasible or significantly higher delays. Compared to the joint worst-case delay minimization benchmarks (Experts~$23$ and~$24$), the agentic AI solution closely approaches their performance, achieving slightly higher computing delay but lower communication delay.

\begin{figure}[t]
\centering
\includegraphics[width=.9\linewidth]{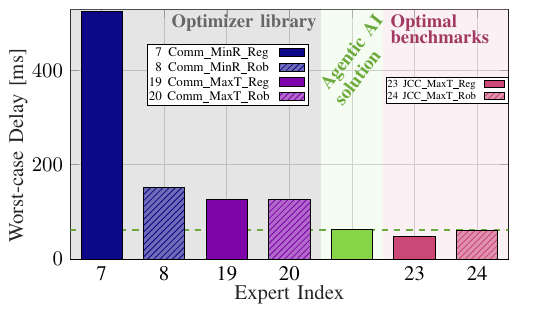}\vspace*{-.2cm}
\caption{Experts, benchmarks, and agentic AI solution space as bar plot, including infeasible results for simulation Set 1.}
\label{fig:set1_bar_single}
\vspace*{-.4cm}
\end{figure}
This observation is further confirmed in Fig.~\ref{fig:set1_bar_single}, which explicitly compares the worst-case joint communication and computing delay. The agentic AI solution consistently outperforms all individual experts while approaching the performance of the optimal joint benchmarks.

\begin{figure}[t]
\centering
\includegraphics[width=.9\linewidth]{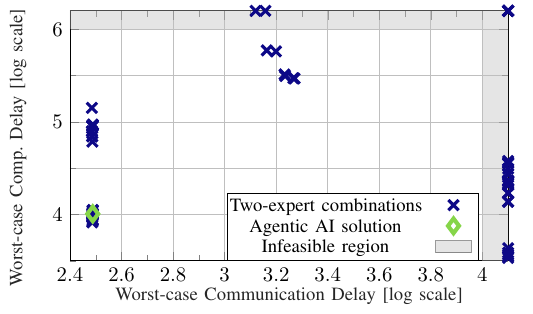}\vspace*{-.2cm}
\caption{Expert combinations and agentic AI solution space as scatter plot, including infeasible results for simulation Set 1.}
\label{fig:set1_scatter_two}
\vspace*{-.4cm}
\end{figure}
\begin{figure}[t]
\centering
\includegraphics[width=.9\linewidth]{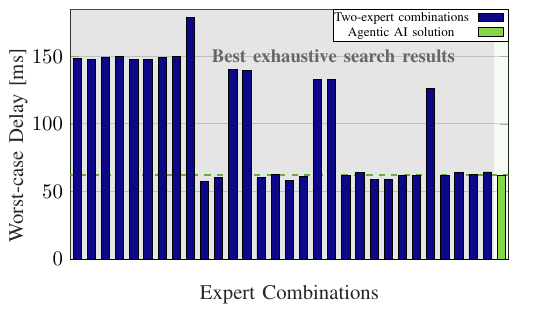}\vspace*{-.2cm}
\caption{Expert combinations and agentic AI solution space as bar plot, including infeasible results for simulation Set 1.}
\label{fig:set1_bar_two}
\vspace*{-.4cm}
\end{figure}
Next, we consider an exhaustive search over all pairwise combinations of experts with equal weighting. The resulting solution space is shown in Fig.~\ref{fig:set1_scatter_two}, particularly showing the worst-case computing delay on the y-axis and worst-case communication delay on the x-axis, including an infeasibility region. While a small number of combinations achieve slightly lower computing delay, the majority of combinations perform worse than the agentic AI solution. Importantly, the agentic AI solution itself corresponds to one of the feasible expert combinations, demonstrating that the LLM-based gate is capable of identifying near-optimal expert compositions without exhaustive enumeration.

Fig.~\ref{fig:set1_bar_two} compares the best-performing expert combinations in terms of the worst-case delay. Fig.~\ref{fig:set1_bar_two} particularly identifies the $30$ best-performing exhaustive search results obtained from Fig.~\ref{fig:set1_scatter_two}, and illustrates their delay performance versus the agentic AI solution. The proposed agentic framework consistently produces feasible solutions that closely approach both optimal benchmark schemes and the best expert combinations. Unlike exhaustive search, whose complexity grows combinatorially with the number of experts, the proposed agentic MoE approach scales efficiently. 

Table~\ref{tab:simsetALL} shows a runtime comparison of all simulation sets. For simulation Set~1, the average end-to-end inference time is $9.363$s, including an average network delay of $1.09$s and expert inference time of $0.177$s. One trial turned infeasible, corresponding to an overall feasibility accuracy of $90\%$. 
\begin{table}[b]
    \centering
    \caption{Runtime comparison across all simulation sets ($10$ independent trials, each), with end-to-end time, average network delay, expert inference time, and overall accuracy.}\label{tab:simsetALL}\vspace*{-.2cm}
    \resizebox{\linewidth}{!}{\begin{tabular}{r l l l l l}
        \cellcolor{black!50}\textcolor{white}{\textbf{Set}} & 
        \cellcolor{black!50}\textcolor{white}{\textbf{}} & 
        \cellcolor{black!50}\textcolor{white}{\textbf{End-to-end}} & 
        \cellcolor{black!50}\textcolor{white}{\textbf{Netw. delay}} & 
        \cellcolor{black!50}\textcolor{white}{\textbf{Inference}} & 
        \cellcolor{black!50}\textcolor{white}{\textbf{Accuracy}}\\
        \cellcolor{black!50}\textcolor{white}{\textbf{1}} & \cellcolor{black!10}avg & \cellcolor{black!10}$9.363$s & \cellcolor{black!10}$1.09$s & \cellcolor{black!10}$0.177$s & \cellcolor{black!10}$90\%$\\
        \cellcolor{black!50}\textcolor{white}{\textbf{2}} & avg & $5.926$s & $1.192$s & $0.176$s & $100\%$\\
        \cellcolor{black!50}\textcolor{white}{\textbf{3}} & \cellcolor{black!10}avg & \cellcolor{black!10}$7.817$s & \cellcolor{black!10}$1.5$s & \cellcolor{black!10}$0.182$s & \cellcolor{black!10}$100\%$\\
    \end{tabular}}
\end{table}

\subsubsection{Simulation Set 2}
\begin{table}[t]
    \centering
    \caption{Description of simulation Set 2.}\label{tab:simset2}\vspace*{-.2cm}
    \resizebox{\linewidth}{!}{\begin{tabular}{r p{6.5cm}}
        \multicolumn{2}{l}{\cellcolor{black!50}\textcolor{white}{\textbf{Framework setup}}}\\
        \cellcolor{black!50}\textcolor{white}{\textbf{Query}} & \cellcolor{black!10}\emph{I want to jointly achieve optimal network throughput and fairness for a robust extended reality scenario}\\
        \cellcolor{black!50}\textcolor{white}{\textbf{Expert library}} & Experts $1$-$12$\\
        \cellcolor{black!50}\textcolor{white}{\textbf{Optimal benchmarks}} & \cellcolor{black!10}Experts $13$-$18$\\
        \multicolumn{2}{l}{\cellcolor{black!50}\textcolor{white}{\textbf{Agentic AI solution}}}\\
        \cellcolor{black!50}\textcolor{white}{\textbf{Interpretation}} & \cellcolor{black!10}\emph{To jointly achieve optimal network throughput and fairness for a robust extended reality scenario, we need to consider both communication and computing aspects under impaired conditions. This requires a combination of experts that focus on robust maximization of sum-rates and robust maximization of minimum rates.}\\
        \cellcolor{black!50}\textcolor{white}{\textbf{Selection}} & $6$ \emph{JCC\_SumR\_Rob} \& $12$ \emph{JCC\_MinR\_Rob}\\
        \cellcolor{black!50}\textcolor{white}{\textbf{Weights}} & \cellcolor{black!10}$\alpha_{6} = \alpha_{12} = 0.5$
    \end{tabular}}
\end{table}
The setup of simulation Set~2 is given in Table~\ref{tab:simset2}. The optimizer library, i.e., a subset of Table~\ref{tab:experts}, consists of sum-rate and min-rate experts trained for communication, computing, or joint communication and computing objectives. In this simulation set, we investigate the generalization and trade-off capabilities of the proposed agentic AI framework when targeting objectives that are not explicitly available as standalone experts in the optimizer library, requiring the framework to combine multiple experts to approximate the desired behavior. To this end, the log-rate experts are excluded from the optimizer library and used solely as optimal benchmarking schemes, thereby providing an upper bound on the achievable performance.

\begin{figure}[t]
\centering
\includegraphics[width=.9\linewidth]{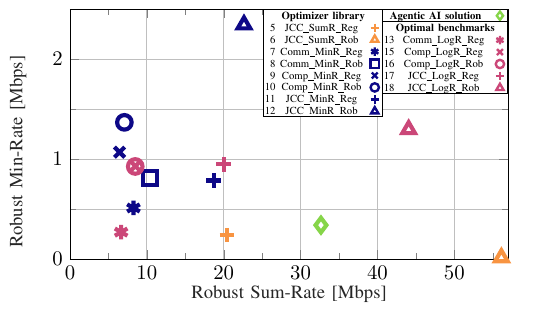}\vspace*{-.2cm}
\caption{Experts, benchmarks, and agentic AI solution space as scatter plot, including infeasible results for simulation Set 2.}
\label{fig:set2_scatter_single}
\vspace*{-.4cm}
\end{figure}
Fig.~\ref{fig:set2_scatter_single} illustrates the achievable robust performance in terms of robust sum-rate and robust min-rate. Specifically, the x-axis shows the robust sum-rate $\lambda_{\text{sumR},n}^\gamma$ in Mbps, defined as $\lambda_{\text{sumR},n} = \sum_{k\in\mathcal{K}} (r_k^\text{tx} + r_k^\text{co})$, while the y-axis shows the robust min-rate $\lambda_{\text{minR},n}^\gamma$ in Mbps, defined as $\lambda_{\text{minR},n} = \min_{k\in\mathcal{K}} (r_k^\text{tx} + r_k^\text{co})$, both evaluated according to \eqref{eq:percentile}. Fig.~\ref{fig:set2_scatter_single} particularly includes all individual library experts, feasible benchmark experts, and the proposed agentic AI solution. The results show that the agentic AI framework outperforms all individual library experts in at least one performance dimension. A similar trend is observed when compared to most benchmark schemes (Experts~$13$ and~$15$-$17$). 

Fig.~\ref{fig:set2_scatter_single} particularly shows that the optimal joint log-rate benchmark (Expert~$18$) achieves higher performance in both robust sum-rate and robust min-rate. This is expected, since Expert~$18$ is trained specifically for robust joint log-rate maximization, which directly balances throughput and fairness. Although such an expert is not available within the optimizer library used by the agentic AI framework, the proposed approach closely approaches its performance, as also illustrated below in Figs.~\ref{fig:set2_single_bar_SR} and~\ref{fig:set2_single_bar_MR}. This demonstrates that the agentic framework is capable of identifying high-quality expert selections that approach the Pareto-optimal performance region.

\begin{figure}[t]
\centering
\includegraphics[width=.9\linewidth]{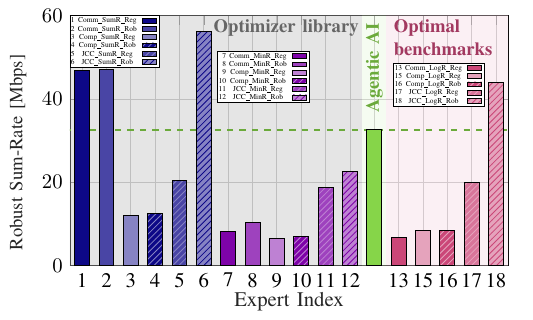}\vspace*{-.2cm}
\caption{Experts, benchmarks, and agentic AI solution space as bar plot showing the sum-rate, including infeasible results for simulation Set 2.}
\label{fig:set2_single_bar_SR}
\vspace*{-.4cm}
\end{figure}
\begin{figure}[t]
\centering
\includegraphics[width=.9\linewidth]{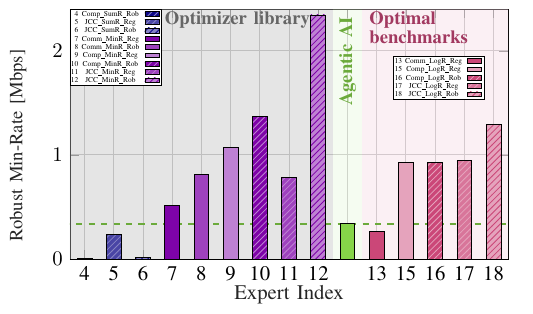}\vspace*{-.2cm}
\caption{Experts, benchmarks, and agentic AI solution space as bar plot showing the min-rate, including infeasible results for simulation Set 2.}
\label{fig:set2_single_bar_MR}
\vspace*{-.4cm}
\end{figure}
These observations are further illustrated in Fig.~\ref{fig:set2_single_bar_SR} and Fig.~\ref{fig:set2_single_bar_MR}, which separately compare the robust sum-rate and robust min-rate for different experts. The agentic AI solution achieves one of the highest robust sum-rate values while maintaining competitive robust min-rate performance, outperforming a good number of experts from within the expert library of Table~\ref{tab:experts}. Again, only the optimal joint benchmark Expert~$18$, consistently exceeds the agentic AI solution in both performance metrics, which is expected given that Expert $18$ is directly trained for robust joint log-rate optimization.

\begin{figure}[t]
\centering
\includegraphics[width=.9\linewidth]{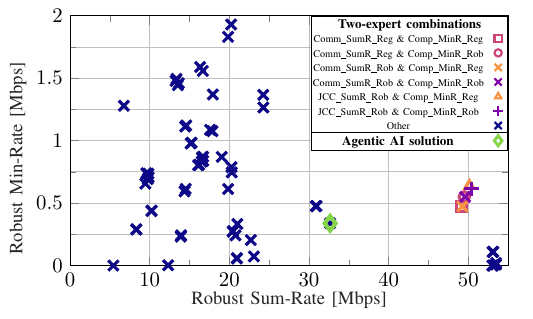}\vspace*{-.2cm}
\caption{Expert combinations and agentic AI solution space as scatter plot, including infeasible results for simulation Set 2.}
\label{fig:set2_scatter_two}
\vspace*{-.4cm}
\end{figure}
Next, we consider exhaustive search over all pairwise combinations of experts with equal weighting. The resulting solution space is shown in Fig.~\ref{fig:set2_scatter_two}. The agentic AI solution itself corresponds to one of the feasible combinations identified by exhaustive search. While a small number of combinations achieve slightly higher performance in both robust sum-rate and robust min-rate, the majority of combinations perform worse than the agentic AI solution. This demonstrates that the proposed agentic framework can identify near-optimal expert compositions with a single inference step, avoiding the combinatorial complexity of exhaustive expert search.

Table~\ref{tab:simsetALL} shows the average end-to-end inference time is $5.962$s, including an average network delay of $1.192$s and expert inference time of $0.176$s. All trials result in feasible solutions, corresponding to an overall feasibility accuracy of $100\%$.

These results confirm that the proposed agentic LLM-enabled MoE framework can efficiently orchestrate specialized optimization experts to achieve robust, high-quality network performance while maintaining scalable inference complexity.

\subsubsection{Simulation Set 3}
\begin{table}[b]
    \centering
    \caption{Description of simulation Set 3.}\label{tab:simset3}\vspace*{-.2cm}
    \resizebox{\linewidth}{!}{\begin{tabular}{r p{6.5cm}}
        \multicolumn{2}{l}{\cellcolor{black!50}\textcolor{white}{\textbf{Framework setup}}}\\
        \cellcolor{black!50}\textcolor{white}{\textbf{Query}} & \cellcolor{black!10}\emph{Computing requirements and channels are subject to estimation errors. I want to serve all users with an equally good delay but also achieve optimal network throughput in terms of data processing and communication}\\
        \cellcolor{black!50}\textcolor{white}{\textbf{Expert library}} & Experts $1$-$30$\\
        \cellcolor{black!50}\textcolor{white}{\textbf{Optimal benchmarks}} & \cellcolor{black!10}----\\
        \multicolumn{2}{l}{\cellcolor{black!50}\textcolor{white}{\textbf{Agentic AI solution}}}\\
        \cellcolor{black!50}\textcolor{white}{\textbf{Interpretation}} & \cellcolor{black!10}\emph{To address your query, we need to focus on both fairness in terms of delay and optimal throughput under conditions where both computing requirements and channels are subject to estimation errors. This involves a combination of experts that handle robust conditions for both delay fairness and throughput optimization.}\\
        \cellcolor{black!50}\textcolor{white}{\textbf{Selection}} & $24$ \emph{JCC\_MaxT\_Rob} \& $6$ \emph{JCC\_SumR\_Rob}\\
        \cellcolor{black!50}\textcolor{white}{\textbf{Weights}} & \cellcolor{black!10}$\alpha_{24} = \alpha_{6} = 0.5$
    \end{tabular}}
\end{table}
The setup of simulation Set~3 is given in Table~\ref{tab:simset3}. The optimizer library consists of the complete set of experts defined in Table~\ref{tab:experts}, covering communication, computing, and joint-domain objectives under both regular and robust operating conditions. In contrast to the previous setups, this scenario provides the full expert library, allowing us to evaluate how effectively the proposed agentic AI framework selects and combines experts to balance competing objectives across a large solution space.

\begin{figure}[t]
\centering
\includegraphics[width=.9\linewidth]{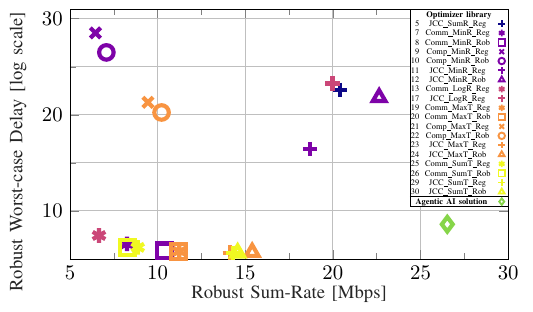}\vspace*{-.2cm}
\caption{Library experts and agentic AI solution space as scatter plot for simulation Set 3.}
\label{fig:set3_scatter_single}
\vspace*{-.4cm}
\end{figure}
Fig.~\ref{fig:set3_scatter_single} illustrates the achievable robust performance in terms of robust sum-rate and robust worst-case delay. The x-axis shows the robust sum-rate $\lambda_{\text{sumR},n}^\gamma$ in Mbps, defined as $\lambda_{\text{sumR},n} = \sum_{k\in\mathcal{K}} (r_k^\text{tx} + r_k^\text{co})$, while the y-axis shows the logarithm of the robust worst-case delay $\log(\lambda_{\text{maxT},n}^\gamma)$, where $\lambda_{\text{maxT},n} = \max_{k\in\mathcal{K}} (t_k^\text{tx} + t_k^\text{co})$, both evaluated according to \eqref{eq:percentile}. Fig.~\ref{fig:set3_scatter_single} includes all individual experts and the agentic AI solution. The results demonstrate that the agentic AI framework outperforms all individual experts in at least one performance dimension, achieving an effective trade-off between robust sum-rate maximization and robust delay minimization. This highlights the ability of the LLM-based gate to combine expert capabilities and balance competing network objectives, thereby enhancing the performance of the proposed agentic AI-based optimization framework.

\begin{figure}[t]
\centering
\includegraphics[width=.9\linewidth]{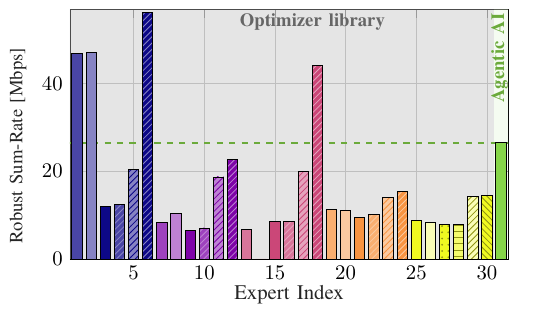}\vspace*{-.2cm}
\caption{Experts and agentic AI solution space as bar plot showing the robust sum-rate for simulation Set 3.}
\label{fig:set3_single_bar_SR}
\vspace*{-.4cm}
\end{figure}
\begin{figure}[t]
\centering
\includegraphics[width=.9\linewidth]{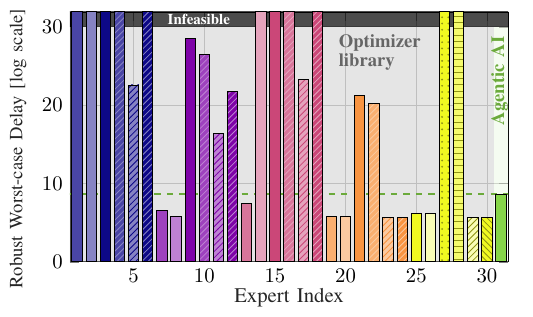}\vspace*{-.2cm}
\caption{Experts and agentic AI solution space as bar plot showing the robust log-max-delay, including infeasible results for simulation Set 3.}
\label{fig:set3_single_bar_MD}
\vspace*{-.4cm}
\end{figure}
\begin{figure}[t]
\centering
\includegraphics[width=.9\linewidth]{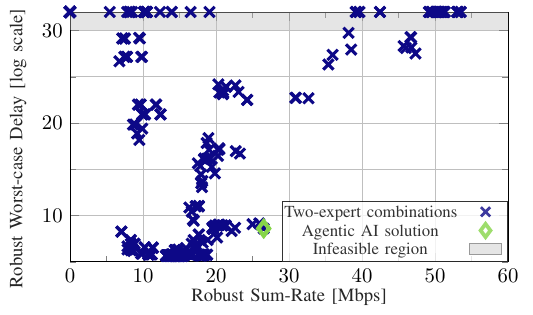}\vspace*{-.2cm}
\caption{Expert combinations and agentic AI solution space as scatter plot for simulation Set 3.}
\label{fig:set3_scatter_two}
\vspace*{-.4cm}
\end{figure}
This observation is further validated in Fig.~\ref{fig:set3_single_bar_SR} and Fig.~\ref{fig:set3_single_bar_MD}, which separately compare robust sum-rate and robust worst-case delay across all $30$ experts. Figs.~\ref{fig:set3_single_bar_SR} and Fig.~\ref{fig:set3_single_bar_MD}, particularly, highlight how four experts achieve higher robust sum-rate than the agentic AI solution, including Expert~$6$, which is selected by the LLM as part of the expert composition. In terms of delay, several experts produce infeasible solutions. Among the feasible experts, the agentic AI solution outperforms eight experts and is outperformed by eleven experts in terms of delay alone, including Expert~$24$, which is also selected by the LLM. However, those experts achieving lower delay typically do so at the expense of significantly reduced sum-rate. In contrast, the agentic AI solution balances both objectives, achieving competitive delay while maintaining strong sum-rate performance. This demonstrates that the proposed framework can combine complementary expert capabilities to obtain a well-balanced trade-off between sum-rate maximization and delay minimization under robust operating conditions.

Next, we consider exhaustive search over all pairwise combinations of experts with equal weighting. The resulting solution space is shown in Fig.~\ref{fig:set3_scatter_two}. The figure particularly highlights how the agentic AI solution is identified among the feasible combinations discovered by exhaustive search, confirming that the LLM-based gate is capable of selecting near-optimal expert compositions without explicitly evaluating the entire combinatorial solution space.

More specifically, across $10$ independent trials, the average end-to-end inference time is $7.817$s, including an average network delay of $1.5$s and expert inference time of $0.182$s, as illustrated in Table~\ref{tab:simsetALL}. All trials, in fact, result in feasible solutions, corresponding to an overall feasibility accuracy of $100\%$.

These results further confirm that the proposed agentic LLM-enabled MoE framework can efficiently orchestrate a large and diverse expert library, achieving robust and balanced performance across competing optimization objectives while maintaining scalable inference complexity.

\subsubsection{Simulation Set 4 -- Accuracy}
To evaluate the expert selection accuracy and robustness of the proposed agentic framework under diverse operator intents, we conduct a larger simulation set consisting of $50$ distinct operator queries. For each query, the expert(s) selected by the LLM-based gate are compared against a reference selection determined by a human expert. The proposed framework achieves an overall accuracy of $84\%$, indicating that in $84\%$ of the cases, the LLM-selected expert composition matches the human reference solution. The remaining $16\%$ correspond to incorrect or partially incorrect selections, such as cases where one expert in a multi-expert composition is correctly selected while another is suboptimal. These errors primarily arise from ambiguous robustness or objective descriptions in the operator query, highlighting the inherent complexity of semantic intent interpretation. For illustration purposes, we also include Table~\ref{tab:simset4}, which presents representative examples of expert selection outcomes. Four examples illustrate successful expert selection, where the LLM-selected experts match the human reference. One failure case is also shown, in which the LLM selected a regular expert instead of a robust expert, indicating incomplete interpretation of the robustness requirement specified in the operator query.

The average end-to-end LLM inference latency in this simulation set is $22.42$\,s, reflecting the combined latency of prompt transmission, LLM reasoning, and expert selection. Overall, these results demonstrate that the proposed agentic LLM-enabled MoE framework can reliably interpret high-level operator intent and select appropriate optimization experts, achieving high semantic selection accuracy while maintaining practical inference latency.

\begin{table}[t]
    \centering
    \caption{Example queries, LLM-selected experts, and accuracy in terms of human-expert solution.}\label{tab:simset4}\vspace*{-.2cm}
    \resizebox{\linewidth}{!}{\begin{tabular}{r p{7.4cm}}
        \multicolumn{2}{l}{\cellcolor{black!50}\textcolor{white}{\textbf{Successful Examples}}}\\
        \cellcolor{black!50}\textcolor{white}{\textbf{Query}} & \cellcolor{black!10}\emph{I have very uncertain network estimates and want to avoid extreme delays for any user. However, throughput should not collapse completely. The network must remain resilient overall.}\\
        \cellcolor{black!50}\textcolor{white}{\textbf{Solution}} & \emph{JCC\_MaxT\_Rob}, $\alpha_{24}=0.6$ \& \emph{JCC\_SumR\_Rob}, $\alpha_{6}=0.4$\\
        \cellcolor{black!50}\textcolor{white}{\textbf{Query}} & \cellcolor{black!10}\emph{I want to ensure that critical communication links achieve guaranteed minimum rates while optimizing the total throughput for the rest under regular conditions.}\\
        \cellcolor{black!50}\textcolor{white}{\textbf{Solution}} & \emph{Comm\_MinR\_Reg}, $\alpha_{7}=0.5$ \& \emph{Comm\_SumR\_Reg}, $\alpha_{1}=0.5$\\
        \cellcolor{black!50}\textcolor{white}{\textbf{Query}} & \cellcolor{black!10}\emph{Channel estimates are almost perfect, but computing workloads fluctuate unpredictably. The main objective is to guarantee fairness in computing performance while maintaining reasonable throughput.}\\
        \cellcolor{black!50}\textcolor{white}{\textbf{Solution}} & \emph{Comp\_MinR\_Rob}, $\alpha_{10}=0.6$ \& \emph{Comm\_SumR\_Reg}, $\alpha_{1}=0.4$\\
        \cellcolor{black!50}\textcolor{white}{\textbf{Query}} & \cellcolor{black!10}\emph{I want to jointly minimize the total communication and computing delay, but the channel estimates are highly uncertain.}\\
        \cellcolor{black!50}\textcolor{white}{\textbf{Solution}} & \emph{JCC\_SumT\_Rob}, $\alpha_{30}=1$\\
        \multicolumn{2}{l}{\cellcolor{black!50}\textcolor{white}{\textbf{Failed Example}}}\\
        \cellcolor{black!50}\textcolor{white}{\textbf{Query}} & \cellcolor{black!10}\emph{The computing must finish within a strict delay limit, but communication links should still achieve a decent sum-rate. Both experience small estimation errors.}\\
        \cellcolor{black!50}\textcolor{white}{\textbf{AI}} & \emph{Comp\_MaxT\_Rob}, $\alpha_{22}=0.6$ \& \emph{Comm\_SumR\_Reg}, $\alpha_{1}=0.4$\\
        \cellcolor{black!50}\textcolor{white}{\textbf{Human}} & \cellcolor{black!10}\emph{Comp\_MaxT\_Rob}, $\alpha_{22}=0.6$ \& \emph{Comm\_SumR\_Rob}, $\alpha_{2}=0.4$
    \end{tabular}}
\end{table}

\section{Conclusion}\label{sec:con}
This paper introduces an agentic AI-based optimization framework that integrates MoE architectures with LLMs for scalable and intent-driven orchestration of specialized optimization agents. By employing an LLM as a semantic gate, the proposed framework enables dynamic expert selection and composition based on high-level operator objectives, bridging the gap between human-readable intent and low-level resource allocation decisions. A general, model-agnostic formulation is presented, followed by a case study on joint communication and computing optimization. A library of specialized deep learning-based experts addresses diverse objectives, including sum-rate, min-rate, log-rate, and delay under regular and robust operating conditions. The LLM-based gate is shown to effectively combine complementary expert capabilities, achieving balanced performance across competing objectives. Numerical simulations also demonstrate that the proposed agentic MoE framework consistently achieves near-optimal performance compared to exhaustive expert search while outperforming individual experts across a wide range of operating scenarios. The proposed framework particularly maintained high feasibility and expert selection accuracy, while avoiding the combinatorial complexity associated with exhaustive expert composition. These results confirm the effectiveness, scalability, and practical feasibility of agentic LLM-driven expert orchestration for AI-native network optimization. Moving forward, it is imperative to address the threat of jail-breaking attacks, where bad actors might trick the orchestrator to manipulate routing policies or degrade network performance. Future investigations should also focus on developing adaptive defense mechanisms, including multi-agent consensus protocols and runtime input sanitization, to secure the proposed agentic AI-based optimization framework's decision-making process without compromising its optimization speed. 

\appendices
\section{Expert Characterization}\label{appdx:A}
\begin{figure*}[t]
\centering
\includegraphics[width=1\linewidth]{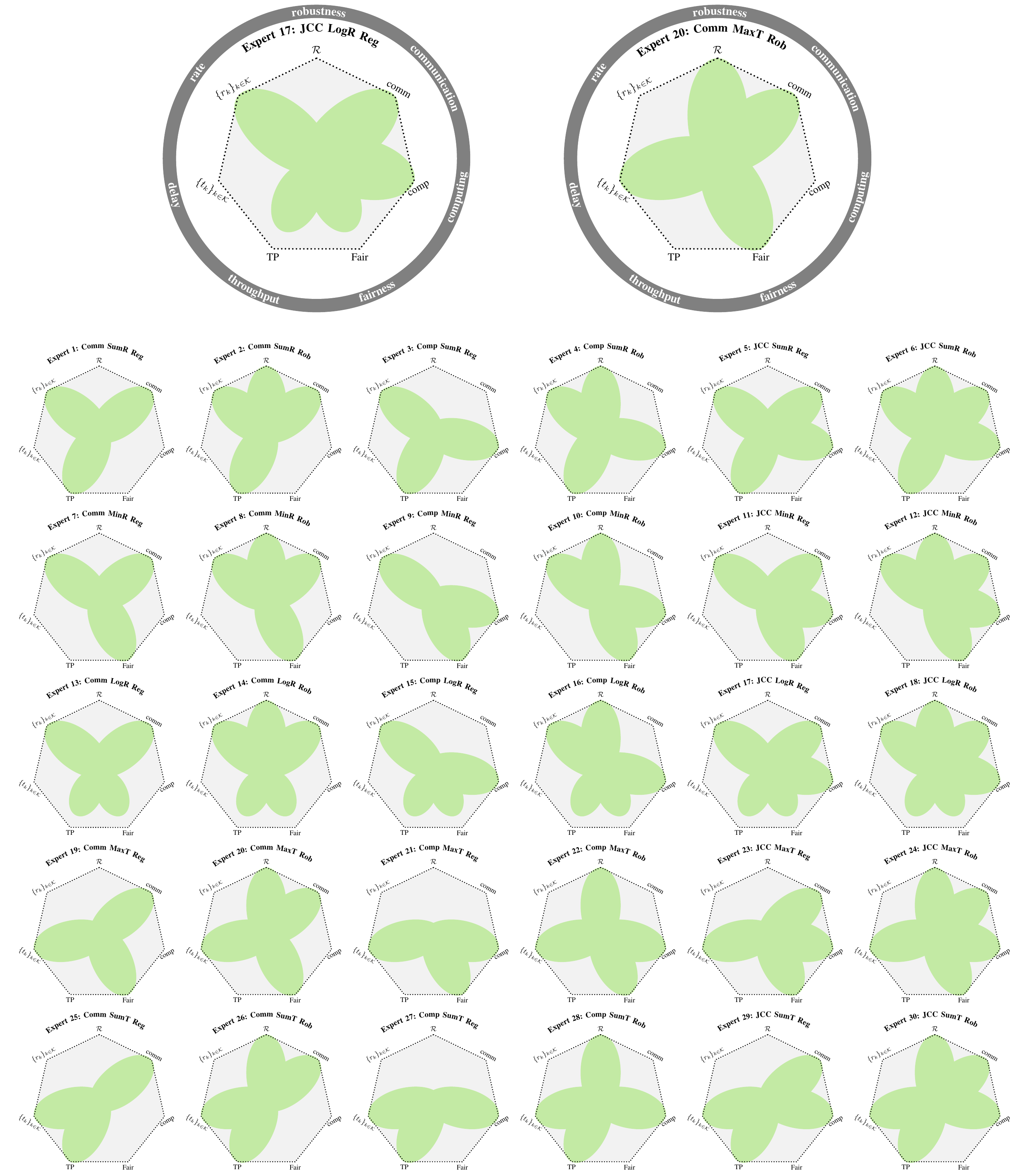}\vspace*{-.2cm}
\caption{Visual characterization of experts from Table~\ref{tab:experts}.}
\label{fig:cake}
\vspace*{-.4cm}
\end{figure*}

Fig.~\ref{fig:cake} provides a qualitative illustration of the distinct characteristics of the $30$ experts presented in Table~\ref{tab:experts} in terms of robustness, communication and computing focus, as well as their emphasis on rate, delay, throughput, and fairness objectives. For illustrative purposes, Fig.~\ref{fig:cake} shows Expert~17 and Expert ~20 on top. Expert~17, also referred to as \emph{JCC\_LogR\_Reg}, maximizes the logarithmic joint communication and computing rate under regular operating conditions, leading to a balanced resource allocation between througput and fairness. Fig.~\ref{fig:cake} particularly highlights the communication (comm) and computing (comp) focus, the throughput and fairness trade-off, and the inclusion of rate metrics only ($\{r_k\}_{k\in\mathcal{K}}$ in the figure). Expert~20, also referred to as \emph{Comm\_MaxT\_Rob}, minimizes the worst-case communication delay under uncertain operating conditions. Fig.~\ref{fig:cake} particularly highlights the robustness ($\mathcal{R}$ in the figure), communication (comm) focus, fairness focus, and the inclusion of delay metrics ($\{t_k\}_{k\in\mathcal{K}}$ in the figure).

\section{System Prompt}\label{appdx:B}
Table~\ref{tab:system_prompt} summarizes the system prompt $\mathcal{Q}^{\text{sys}}$ employed in the case study. The prompt is structured into three components: the \emph{general setup}, the \emph{expert setup}, and the \emph{tool setup}. The general setup describes the overall framework, the network model, and the role of the LLM as a gating mechanism. The expert setup introduces each expert from Table~\ref{tab:experts} using human-readable descriptions. Finally, the tool setup defines the available tool calls, distinguishing between single-expert and multi-expert selection and specifying the corresponding rules for expert weighting and selection.
\begin{table}[t]
\caption{Excerpt of the system prompt $\mathcal{Q}^{\text{sys}}$ used for the LLM-enabled gate network.}
\label{tab:system_prompt}
\centering
\small
\begin{tabular}{r p{0.9\linewidth}}
        \multicolumn{2}{l}{\cellcolor{black!50}\textcolor{white}{\textbf{General setup}}}\\
        \cellcolor{black!50}\textcolor{white}{} & \emph{You are a helpful assistant that receive requests in natural language and then, based on the provided context you must choose one or a combination of suitable expert tools which can resolve the query asked. You operate as an intelligent assistant embedded within a wireless network operator environment. You are functioning as a router/gate network that receives a query or a question from the network operator and your task is to route the question to an optimization expert.}\\
        \multicolumn{2}{l}{\cellcolor{black!50}\textcolor{white}{\textbf{Expert setup}}}\\
        \cellcolor{black!50}\textcolor{white}{} & \cellcolor{black!10}\emph{You have 30 experts that can be used to resolve queries either by their own or on combinations. This depends on the query if the requested information need to be a combination of the results from several experts or one expert can fully address the query. This can be solely determined by the description of each expert area of expertise. Here is a detailed description of the available experts and their area of specialization:}\\
        \cellcolor{black!50}\textcolor{white}{} & \emph{1) Comm\_SumR\_Reg: Expert for maximization of sum-communication-rates of all users in network. Communication throughput focused. Optimized solution accounts for regular conditions, with accurate channel estimations and perfectly known channel state information.}\\
        \cellcolor{black!50}\textcolor{white}{} & \cellcolor{black!10}\emph{[\,\ldots\,]}\\
        \multicolumn{2}{l}{\cellcolor{black!50}\textcolor{white}{\textbf{Tool setup}}}\\
        \cellcolor{black!50}\textcolor{white}{} & \emph{The available tools for the router are listed below:}\\
        \cellcolor{black!50}\textcolor{white}{} & \cellcolor{black!10}\emph{infer\_expert\_with\_params: This function take a string as expert name and based on this input it infers the correct expert with given parameters and return a tuple of three parameters. The expert name is a string that can be chosen of a set of specific available expert names.}\\
        \cellcolor{black!50}\textcolor{white}{} & \emph{infer\_two\_weighted\_experts\_with\_params: This function take two strings as expert names and two numeric parameters alpha\_1 and alpha\_2 where alpha\_1 + alpha\_2 = 1. Based on these input parameters the function combine the inference results from the two given experts with the weighting parameters alpha\_1 and alpha\_2 and return a tuple of three parameters. The expert name is a string that can be chosen of a set of specific available expert names.}\\
        \cellcolor{black!50}\textcolor{white}{} & \cellcolor{black!10}\emph{[\,\ldots\,]}
    \end{tabular}
\end{table}

\bibliographystyle{IEEEtran}
\bibliography{bibliography}

\end{document}